\documentclass[10pt,twocolumn,letterpaper]{article}

\usepackage{cvpr}               
\usepackage{bbm}

\usepackage{booktabs}
\usepackage{multirow}
\usepackage{rotating}
\usepackage{adjustbox}
\def\eg{\emph{e.g}\onedot} 
\def\ie{\emph{i.e}\onedot} 
\usepackage{mathtools}
\usepackage{textcomp}
\usepackage{gensymb}
\usepackage{enumitem}
\usepackage{color, colortbl}
\usepackage{makecell}
\usepackage[dvipsnames]{xcolor}

\usepackage[hang]{footmisc}
\definecolor{cvprblue}{rgb}{0.21,0.49,0.74}
\usepackage[pagebackref,breaklinks,colorlinks,allcolors=cvprblue]{hyperref}
\usepackage[capitalize]{cleveref}
\crefname{section}{Sec.}{Secs.}
\Crefname{section}{Section}{Sections}
\Crefname{table}{Table}{Tables}
\crefname{table}{Tab.}{Tabs.}
\newcommand{\pub}[1]{\color{gray}{\scriptsize{[{#1}]}}}
\newcommand\ours{SAMWISE}

\newcommand\sam{\textsc{SAM2}}
\newcommand{\myparagraph}[1]{\vspace{4pt}\noindent\textbf{#1.}}
\definecolor{mylightgray}{gray}{0.92}

\title{SAMWISE: Infusing Wisdom in SAM2 for Text-Driven Video Segmentation}

\author{Claudia Cuttano$^{1}$ \quad Gabriele Trivigno$^{1}$
\quad
Gabriele Rosi$^{1, 2}$
\quad
Carlo Masone$^{1,2}$
\quad
Giuseppe Averta$^{1,2}$\\
$^{1}$ Politecnico di Torino
$^{2}$ Focoos AI \\
{\tt\small \{name.surname\}@polito.it \, \{name.surname\}@focoos.ai}
}

\begin{document}

\twocolumn[{%
\renewcommand\twocolumn[1][]{#1}
\maketitle
\begin{center}
    \captionsetup{type=figure}
    \includegraphics[width=\linewidth]{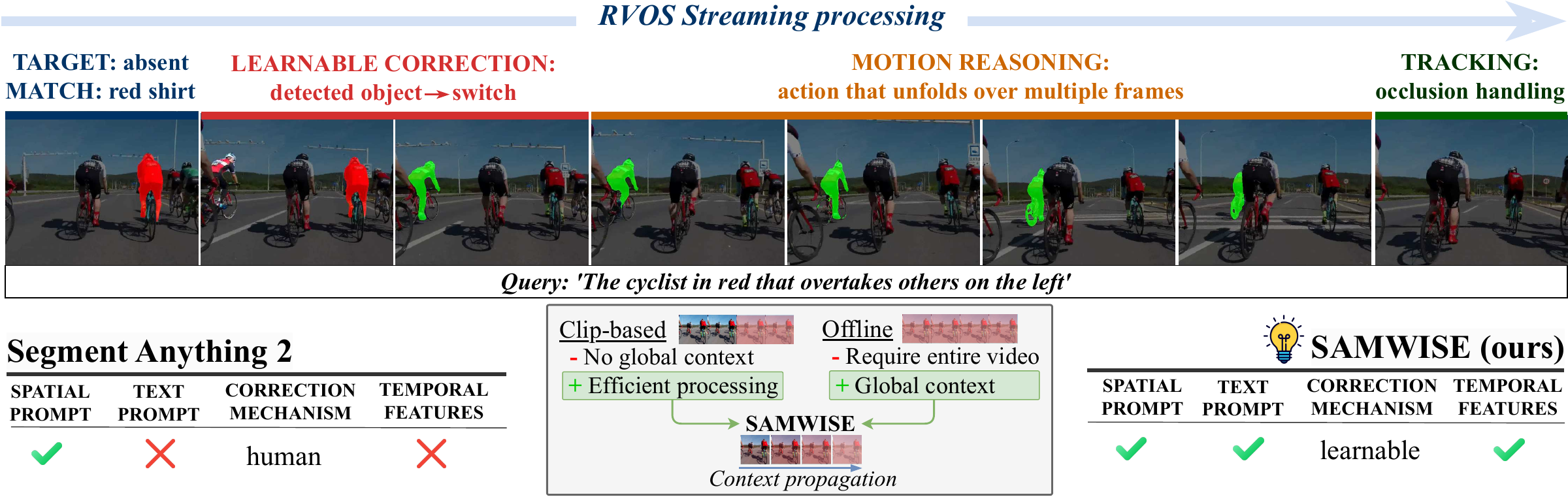}
    \vspace{-0.1cm}
    \captionof{figure}
  {\textbf{SAMWISE}. 
    Our approach infuses knowledge about natural language in the Segment-Anything 2 model, adding explicit temporal cues in the feature extraction for the task of streaming-based Referring Video Segmentation (RVOS). We use a learnable mechanism to  mitigate the so-called \textit{tracking} bias, \ie 
    \sam{} tendency to overlook a correct object once it becomes identifiable, due to its ongoing tracking of a different object.
    Our design enables effective streaming processing for RVOS, exploiting the memory from previous frames to propagate past context.
    The figure shows an example where the target object is not present in the first frame, leading \sam{} to start tracking the wrong one.
    Afterwards, when the correct object appears, our learnable correction mechanisms guides \sam{} to switch its tracking focus. By adding in its features the notion of temporal evolution, the model is able to recognize that the new object is more aligned with the provided textual query. Finally, we exploit \sam{} tracking skills and robustness to occlusions to keep following the object. 
    }
    \label{fig:teaser}
\end{center}
}]

\begin{abstract}
Referring Video Object Segmentation (RVOS) relies on natural language expressions to segment an object in a video clip.
Existing methods restrict reasoning either to independent short clips, losing global context, or process the entire video offline, impairing their application in a streaming fashion. 
In this work, we aim to surpass these limitations and design an RVOS method capable of effectively operating in streaming-like scenarios while retaining contextual information from past frames. 
We build upon the Segment-Anything 2 (SAM2) model, that provides robust segmentation and tracking capabilities and is naturally suited for streaming processing. We make \sam{} wiser, by empowering it with natural language understanding and explicit temporal modeling  at the feature extraction stage, without fine-tuning its weights, and without outsourcing modality interaction to external models. To this end, we introduce a novel adapter module that injects temporal information and multi-modal cues in the feature extraction process. We further reveal the phenomenon of tracking bias in \sam{} and propose a learnable module to adjust its tracking focus when the current frame features suggest a new object more aligned with the caption. 
Our proposed method, \ours, achieves state-of-the-art across various benchmarks, by adding a negligible overhead of less than 5 M parameters.
Code is available at {\small{\url{https://github.com/ClaudiaCuttano/SAMWISE}}}.
\end{abstract}    
\vspace{-0.15cm}
\section{Introduction}
\label{sec:intro}

Referring video segmentation (RVOS) ~\cite{Seo_2020_urvos, gavrilyuk_2018_actor, khoreva_2019_davis, ye_2019_cmsa, liu_2021_cmpc, wu_2022_language} aims at segmenting and tracking specific objects of interest within video content, guided by natural language expressions \cite{han_2023_html, luo_2024_soc, botach_2022_mttr}.
%is a blooming research area that lies at the intersection of computer vision and natural language processing. It 
Existing RVOS methods are mostly based on a \emph{divide and conquer} paradigm, where the video is divided into shorter clips that are processed independently \cite{wu_2022_language, botach_2022_mttr, tang_2023_tempcd}.
However, as demonstrated by MeViS~\cite{ding_2023_mevis}, this solution fails in examples that require taking into account long-term motion and global context.
As a workaround to handle this challenge, the state-of-the-art method \cite{he_2024_decoupling} processes the entire video in an \textit{offline} fashion, first modeling trajectories of all instances throughout the entire clip and then selecting the most appropriate one. 
Albeit effective, this approach is not applicable when the model has access only to a portion of the video, for example when the data at inference time are presented in a streaming fashion or due to limitations in the computational resources. The trade-off of these two paradigms is schematized in \cref{fig:teaser}. To this end, OnlineRefer \cite{wu_2023_online} introduced a context propagation scheme for \textit{online} RVOS but relies solely on past context from a single frame, limiting its ability to capture long-term dependencies.
In this work, we investigate how to exploit the memory from past frames to design an RVOS method capable of retaining global context while operating within a streaming paradigm, \ie, without requiring access to the whole video at once. 
This idea is inspired by the recent release of Segment-Anything 2 (\sam{}) \cite{ravi_2024_sam2}, a foundational model that has shown impressive capabilities in various Video Segmentation tasks thanks to a memory bank that allows to leverage long-range past information. Since \sam{} operates in a streaming fashion, extending this method to enable context-aware streaming processing in RVOS would appear a natural step.
However, this entails some non-trivial challenges:

\noindent
\textbf{i) Text understanding.} \sam{} original design accounts only for \textit{spatial} prompts (e.g. points) and lacks mechanisms to interpret \textit{semantic} prompts like text, which require reasoning over visual and textual modalities. 
While we are the first to address the challenge of adding textual prompts to \sam{}, previous methods have explored this problem for SAM-1 at image-level. 
These solutions \cite{zhang_2024_evf, lai2024lisa} delegate visual-textual interaction to an off-the-shelf large VLM ( like BEIT-3 \cite{wang_2022_beit}, LLaVa \cite{liu_2024_llava}), which generates a multi-modal embedding that is used to prompt SAM-1.

\noindent
\textbf{ii) Temporal modeling.}  To segment the referred object throughout the video, it must be first \textit{recognized} and then \textit{tracked}. While the latter requires matching objects visual appearance across adjacent frame, the recognition problem entails modeling temporal evolution to reason over actions that unfold over multiple frames. However, \sam{} extracts frame features independently, lacking such reasoning. 

\noindent
\textbf{iii) Tracking bias.}
In RVOS, the target object might be unrecognizable during certain time intervals, due to occlusions, presence of multiple instances or forthcoming actions, as in the first frames of \cref{fig:teaser}. In such cases, \sam{} may start tracking an incorrect object that partially matches the textual prompt, and persist in following it, leading to what we denote as \textit{tracking bias}. While \sam{}{} original design allows for a user to manually correct the prediction by providing a new prompt, such a strategy is not applicable in tasks without a human-in-the-loop like RVOS.

\noindent
In this work, we aim at making \sam{} \textit{wiser}, by addressing these limitations without fine-tuning \sam{} weights, thereby preserving its original capabilities, and without outsourcing modality interaction to external, heavy models.
To overcome challenges $i)$ and $ii)$, we design a learnable Adapter \cite{houlsby_2019_adapter} module, named Cross-Modal Temporal Adapter (CMT), with two key principles in mind: a) enabling mutual interaction between visual and linguistic modalities;  and b) encoding temporal cues into visual features. Then, to generate a prompt, we follow  \cite{zhu_2023_minigpt, liu_2024_llava} and employ a learnable MLP to project the sentence embedding for the \sam{} Mask Decoder, which then outputs the final segmentation mask.
In this way, we can exploit \sam{} tracking capability to segment an object given a textual query across the video. Finally, to mitigate the \textit{tracking bias} problem $iii)$, we introduce a lightweight Conditional Memory Encoder (CME) which detects when a candidate object, aligned with the text, appears in the frame, thus enabling \sam{} to dynamically refocus its tracking to the correct object as it becomes distinguishable. 

\noindent
Summarizing, this paper contributes with the following:
\begin{itemize}
    \item We present \ours{}, the first method that integrates natural language knowledge into \sam{} in an end-to-end solution tailored to address the challenges of RVOS. We introduce a novel adapter, namely Cross Modal Temporal (CMT) Adapter, which purposefully models temporal evolution and multi-modal interaction;
    \item We provide insight into the functioning of \sam{}, highlighting the phenomenon of \textit{tracking bias}, and introduce a learnable module (Conditional Memory Encoder) to adjust tracking based on new information;
    \item Our methods achieves state-of-the-art results both on traditional RVOS benchmarks (Ref-Youtube-VOS \cite{Seo_2020_urvos}, Ref-DAVIS \cite{khoreva_2019_davis}), as well as the more challenging MeViS \cite{ding_2023_mevis}, without compromising \sam{} capabilities and adding less than $5 M$ learnable parameters.
\end{itemize}
\section{Related works}
\label{sec:related}
\begin{figure*}[t]
    \centering
    \includegraphics[width=0.99\linewidth]{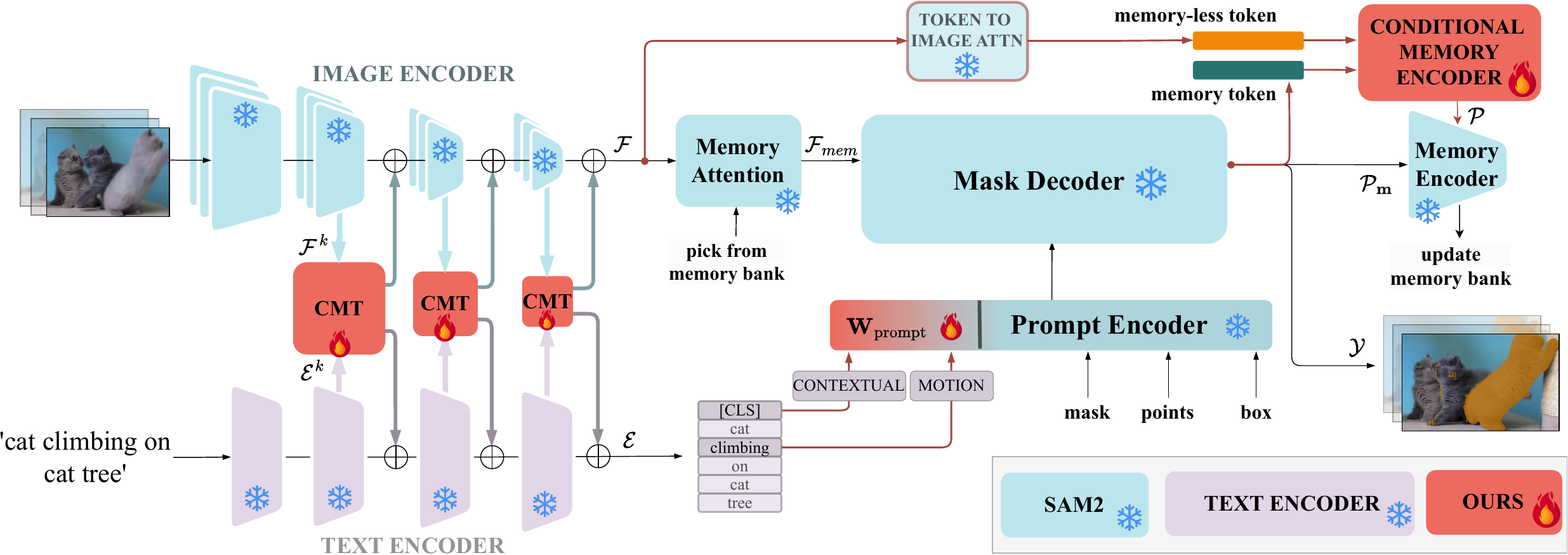}
    \vspace{0.2cm}
    \caption{\textbf{Overview of \ours.}
    We build on a frozen \sam{} and a frozen Text Encoder to segment images in video given a textual description. 
    We incorporate the Cross-Modal Temporal Adapter (CMT) into the text and visual encoders at every intermediate layer $k$ to model temporal dynamics within visual features while contaminating each modality with the other. Then, we extract the \texttt{[CLS]} and verb embeddings, namely Contextual and Motion prompts, from the adapted textual features and project them through a learnable MLP. The final embedding is used to prompt the Mask Decoder, which outputs the segmentation mask. Finally, the Conditional Memory Encoder detects when a new candidate object, aligned with the caption, appears in the frame, enabling SAM2 to dynamically refocus its tracking. }
     %\vspace{-15pt}
    \label{fig:method}
\end{figure*}

\myparagraph{Referring Video Segmentation}
In RVOS, the goal is to segment an object in a clip described with natural language queries ~\cite{liu_2021_cmpc,ding_2021_pminet, khoreva_2019_davis, Seo_2020_urvos, wu_2022_language, he_2024_decoupling}. 
Earlier works proposed adapted image-based methods \cite{gavrilyuk_2018_actor, khoreva_2019_davis, ye_2019_cmsa, bellver_2023_refvos}, or used a spatio-temporal memory to attend to masks of previous frames \cite{Seo_2020_urvos, oh_2019_memory}. 
Subsequent works employ a DETR-like \cite{carion_2020_end} structure to process multiple frames and text embeddings \cite{wu_2022_language,botach_2022_mttr,miao_2023_sgmg,han_2023_html}. 
All these methods process short clips independently, thus losing global context.

Recently, \cite{ding_2023_mevis} showed how traditional RVOS benchmarks lack challenging captions that require to disambiguate between instances and their actions, as well as occlusions and dynamic queries, highlighting how they could be solved even with image-based methods. The MeViS dataset \cite{ding_2023_mevis} targets these scenarios, with challenging examples that previous image or clip-based methods fail to address.
To this end, a few works proposed \textit{offline} methods to explicitly model multiple object trajectories \cite{luo_2024_soc, he_2024_decoupling}, with the latter representing the state-of-the-art on MeViS.
Concurrently, OnlineRefer \cite{wu_2023_online} proposed a first attempt towards an \textit{online} RVOS setting, with a query propagation scheme. However, its effectiveness is limited as predictions are based on a single frame.
Our method builds on this paradigm by leveraging \sam{}{} memory bank to encode long-range past context.

\myparagraph{Text-prompted Segment-Anything}
Recent works have provided solutions to adapt SAM-1 for text-prompted segmentation. Grounded SAM \cite{ren_2024_groundedsam} employs a two step pipeline where GroundingDINO \cite{liu_2023_grounding} generates bounding boxes for SAM-1 to produce segmentation masks. Applying such pipeline in RVOS is problematic, as potential errors in the first frame are propagated throughout the whole video.
To directly prompt SAM-1, RefSAM \cite{li_2024_refsam} exploits a projection layer to map the textual embedding into the prompt space, while \cite{lai2024lisa, yan_2024_visa, bai_2024_onetoken} resort to large off-the-shelf VLM to generate a multi-modal embedding that is used to prompt SAM-1. Both solutions finetune the Mask Decoder, thereby compromising its capabilities on its original task. 
In contrast, our work is the first to propose an end-to-end model that incorporates textual knowledge within \sam{} without fine-tuning nor relying on external models.

\myparagraph{Pre-Trained Knowledge Transfer}
In recent years, the release of powerful pretrained models has sparked interest in the question of how to extend their skills to novel tasks, as full fine-tuning becomes increasingly impractical with growing model sizes \cite{peters_2019_tune, jin_2024_mv}.
A powerful strategy to address this problem relies on using Adapters \cite{houlsby_2019_adapter}, small trainable modules that enable efficient adaptation of pre-trained models.
Following this paradigm, recent studies have explored adapting CLIP \cite{radford_2021_clip} for downstream tasks. At the image level, \cite{xu_2023_bridging} inserts Transformer Decoder blocks within CLIP encoders, which entail costly Self-Attentions on all tokens. For video tasks, \cite{wang_2024_multimodal} places independent adapter modules within each encoder, whereas \cite{lu_2023_uniadapter, yang_2024_mma, jin_2024_mv, jiang_2022_cross} rely on a weight-sharing mechanism to project both modalities in a shared sub-space.
Nevertheless, as features of each modality are independently extracted, none of these adapters allows explicit feature interaction, unlike our CMT, which also incorporates temporal modeling. Lastly, all these works start from a model that already includes a text encoder (CLIP), whereas ours is the first to propose an adapter for the Segment-Anything 2 model to add textual understanding, achieving robust performances while introducing less than 5 M parameters.

\section{\ours{}}
\label{sec:method_ours}

\myparagraph{Problem setting}
Given an input video $ \mathcal{V}$ $ = \{I_{t}\}_{t=1}^{T_V}$ with $T_V$ frames and a referring expression, we aim to predict a set of binary masks $S = \{s_{t}\}_{t=1}^{T_V}$, $s_{t} \in \mathbb{R}^{H \times W}$ of the referred object.
We tokenize the textual query in a set of $L$ words, $E = \{ e_{l} \}_{l=1}^{L}$, and add a global sentence representation token \texttt{[CLS]}.
The tokens are then processed using a frozen text encoder to extract language features $\mathcal{E}$ $\in \mathbb{R}^{L \times C_t}$.  
We process videos in a streaming fashion, collecting clips of $T$ frames as they are available. Throughout the rest of the section, we use $T$ to indicate clip length.

\myparagraph{Overview}
We first provide a brief discussion of the \sam{} model (\cref{sec:background}). We then outline the pipeline of our proposed \ours{}, starting from the prompting strategy in \cref{sec:prompting}. In \cref{sec:cmta}, we detail our novel Cross-Modal Temporal Adapter.
Lastly, in \cref{sec:cme}, we discuss our learnable correction strategy, named Conditional Memory Encoder, to address the issue of \textit{tracking bias}.

\subsection{Background: Segment-Anything}
\label{sec:background}
The Segment-Anything Model 2 (SAM2) builds upon SAM-1 \cite{kirillov_2023_sam} to tackle the task of Promptable Video Object Segmentation, \ie, tracking an object in a video given a textual prompt. Following SAM-1, it consists of an \emph{image encoder}, a \emph{Prompt Encoder} and a \emph{Mask Decoder}, which combines the image and prompt embeddings to predict segmentation masks. To enable video processing, \sam{} comes with a few modifications: $i)$ the original ViT backbone is replaced by Hiera \cite{ryali_2023_hiera}, roughly 3 times faster, which processes frames independently to provide hierarchical visual features. Hereinafter, we refer to them as \textit{memory-less features} $\mathcal{F}$; $ii)$ frame embeddings are not directly fed to the Mask Decoder, but they are first \textit{conditioned} on memories of past predictions from a \emph{Memory Bank}. We refer to these conditioned features as \textit{memory features} $\mathcal{F}_{mem}$. Lastly, $iii)$ once the mask for the current frame is predicted, the \emph{Memory Encoder} updates the Memory Bank. By design, \sam{} handles video frames as they become available, progressively encoding the past in its Memory Bank. We argue that this streaming approach is especially valuable in RVOS, enabling reasoning over a wide temporal horizon.

\begin{figure}[t]
    \centering
   \includegraphics[width=1.0\linewidth]{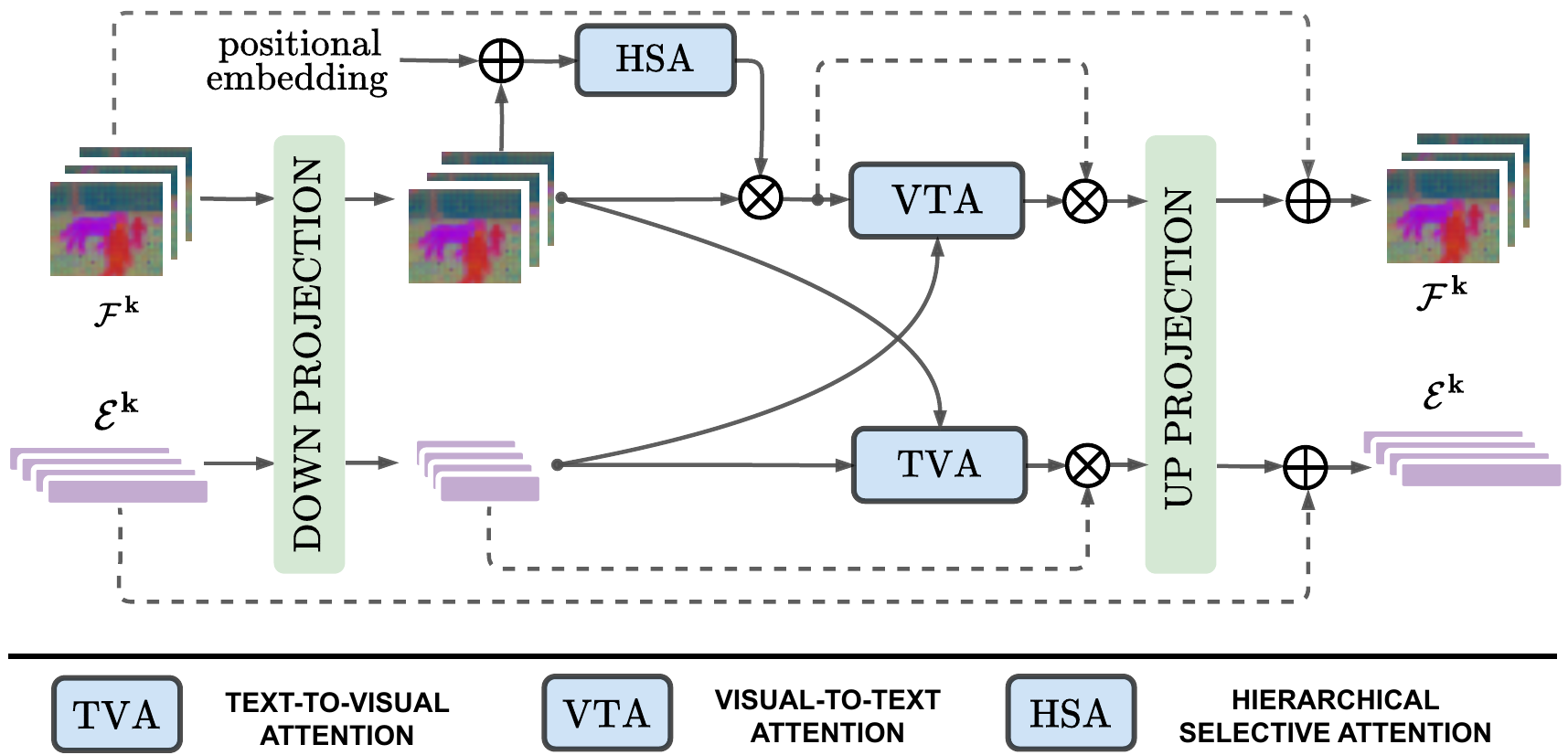}
    \vspace{-0.1cm}
    \caption{Architecture of our \textbf{Cross Modal Temporal} (CMT) Adapter, made up of Hierarchical Selective Attention (HSA) to model temporal cues, a Visual-to-Text Attention (VTA) and Text-to-Visual Attention (VTA) modules.} 
   
    \label{fig:cmt}
    %\vspace{-0.3cm}
\end{figure}

\subsection{Prompting SAM2}
\label{sec:prompting}
To guide the \sam{} decoder, we use a Contextual Prompt, $\mathcal{E}_C \in \mathbb{R}^{1 \times C_t}$, which encodes the high-level semantic information for the given text query, emphasizing the essential aspects of the query while downplaying less relevant elements. To this end, we employ the \texttt{[CLS]} embedding of text features, $\mathcal{E}$. Furthermore, we also introduce a second prompt, the Motion Prompt $\mathcal{E}_M \in \mathbb{R}^{1 \times C_t}$, which captures action-related cues by using verb embeddings from $\mathcal{E}$.
These prompts are concatenated and projected through a learnable three-layer MLP: 
\begin{equation} 
\label{eq:prompt}
\rho = \boldsymbol{\mathrm{W}}_{\text{prompt}}(\texttt{CAT}[\mathcal{E}_C, \mathcal{E}_M]).
\end{equation}
In this way, the provided prompts encode both subject-related and motion-based information. 
Given that in our task the textual prompt is not referred a-priori to any particular frame, we prompt SAM2 at each frame, so that the model has to balance the influence of tracking while also considering the content of each frame. We discuss more in depth this aspect in \cref{sec:cme}.

\subsection{Cross-Modal Temporal Adapter}
\label{sec:cmta}
An adapter consists of a linear down-projection ($ \boldsymbol{\mathrm{W}}_{\text{down}}$) to a bottleneck dimensionality, followed by an up-projection back ($\boldsymbol{\mathrm{W}}_{\text{up}}$) in the original space, separated by a non-linear activation function $\sigma$. Formally, given an input feature $\mathbf{x} \in \mathbb{R}^{1 \times d}$, the adapter function is defined as:
\begin{equation}
    Adapter(\mathbf{x}) = \mathbf{x} + \sigma(\mathbf{x} \boldsymbol{\mathrm{W}}_{\text{down}})\boldsymbol{\mathrm{W}}_{\text{up}}
\end{equation}
We build on this popular Adapter framework \cite{houlsby_2019_adapter} and propose a novel Cross-Modal Temporal Adapter (CMT) (see \cref{fig:cmt}) which models temporal dynamics within visual features while enriching each modality with the other. 
Formally, given the visual features in a clip
$\mathcal{F}^k \in \mathbb{R}^{T \times H_k \times W_k \times C_k}$ and the textual features $\mathcal{E
}^k \in \mathbb{R}^{L \times C}$ extracted at layer $k$ of the image and text encoders, respectively, the CMT can be formulated as:
\begin{equation}
\begin{array}{c}
\begin{aligned}
Adapter(\mathcal{F}^k) = &  \mathcal{F}^k + h(\mathcal{F}^k \boldsymbol{\mathrm{W}}_{\text{down,v}}, \mathcal{E}^k \boldsymbol{\mathrm{W}}_{\text{down, t}})\boldsymbol{\mathrm{W}}_{\text{up,v}} \\
Adapter(\mathcal{E}^k) = & \mathcal{E}^k + h(\mathcal{E}^k \boldsymbol{\mathrm{W}}_{\text{down,t}}, \mathcal{F}^k \boldsymbol{\mathrm{W}}_{\text{down,v}})\boldsymbol{\mathrm{W}}_{\text{up,t}}
\end{aligned}
\end{array}
\end{equation}

\noindent
where $\boldsymbol{\mathrm{W}}_{\text{down,v}}$, $\boldsymbol{\mathrm{W}}_{\text{down,t}}$, $\boldsymbol{\mathrm{W}}_{\text{up,v}}$,$\boldsymbol{\mathrm{W}}_{\text{up,t}}$
are modality specific down- and up-projections weights and $h$ is our proposed adapter function. 
The adapter output is summed with the original features, allowing the model to retain the original encoding while incorporating temporal and cross-modal reasoning.
We integrate the Cross-Modal Temporal Adapter (CMT) into the frozen text and visual encoders at every intermediate layer $k$.
In the following paragraphs we detail the temporal and cross-modal adaptation functions, which are tightly coupled in our Adapter module.

\begin{figure}[t]
    \centering
   \includegraphics[width=1.0\linewidth]{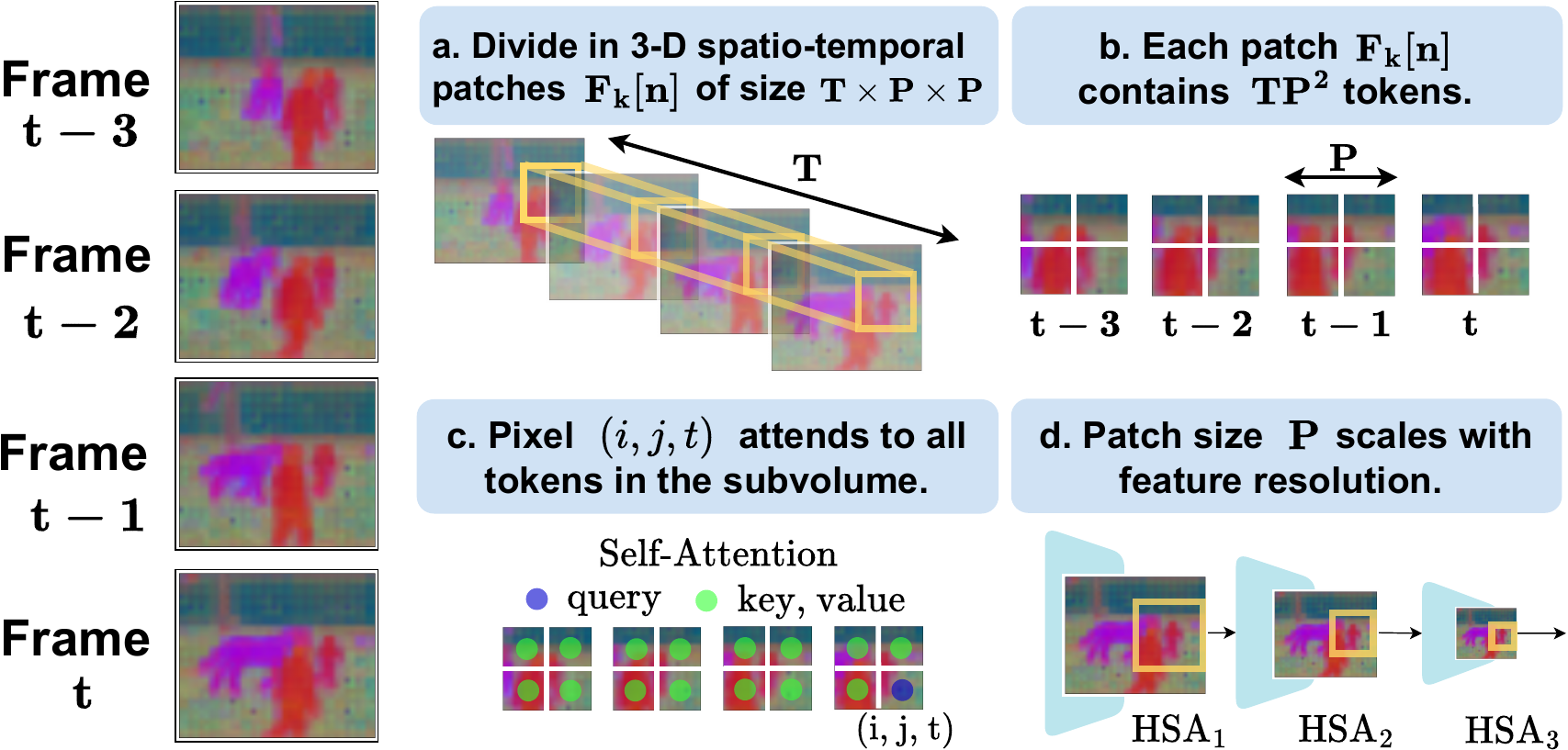}
   \vspace{-0.1cm}
    \caption{Scheme of our \textbf{Hierarchical Selective Attention} (HSA), modeling temporal evolution of features in our adapter.} \vspace{0.1cm}
    \label{fig:hsa}
    %\vspace{-0.2cm}
\end{figure}

\myparagraph{Temporal Adaptation}
Our approach aims to embed motion cues directly into the frame-level features of \sam{}. 
Previous works based on Adapters either perform self-attention (SA) over all tokens in a clip \cite{jin_2024_mv}, which is costly, or restrict the attention to the temporal axis for each pixel \cite{Liu_2024_btadapter, liu_2023_revisiting}.
We observe that, within a video, object motion across adjacent frames typically spans a localized region of the image \cite{patrick_2021_keeping}.
Consequently, a given element of the feature map primarily benefits from interactions with its spatial and temporal neighbors, rather than requiring long-range connections across the entire feature map.
Building on this intuition, we introduce a Hierarchical Selective Attention (HSA) mechanism, illustrated in \cref{fig:hsa}. By modeling interactions among spatially and temporally proximal regions, HSA reduces unnecessary computations while capturing motion-based context.

Formally, at layer $k$, given the set of feature maps for a $T$-frames clip:
$\mathcal{F}^k \in \mathbb{R}^{T \times H_k \times W_k \times C_k}$, we decompose this feature volume into non-overlapping, 3-D spatio-temporal patches of size $T \times P \times P$, obtaining $N=H_kW_k / P^2$ sub-volumes.
These sub-volumes, considered pixelwise, can be represented as a set of tokens 
$F^k[n] = \left\{ x^{k, n}_{i,j,t} \in \mathbb{R}^{C_k} : i \in 1,\ldots,P, j \in 1,\ldots,P, t \in 1,\ldots,T \right\}$. 
To encode spatio-temporal positioning, to each vector we add a spatial ($e[i, j]$) and a temporal ($e[t]$) sinusoidal positional embeddings, in 2-D and 1-D formats, respectively.
%Each vector is treated as a token, to which we sum both a spatial ($e[i, j]$) and temporal ($e[t]$) sinusoidal positional embedding, respectively 2-D and 1-D, to encode their relative spatio-temporal location. 
Specifically: $ x^{k, n}_{i,j,t} = x^{k, n}_{i,j,t} + e[i, j] + e[t]$.
Each sub-volume contains $M=P^2*T$ tokens, on which we perform self-attention as follows:
\begin{equation}
    {\bf x}^{k, n}_{i, j, t} := SA\left( \left\{ {\bf x}^{k, n}_{i', j', t'}  \right\}_{\begin{subarray}{l}i'=1..P\\j'=1..P\\t'=1..T \end{subarray}} 
    %\right]
    \right).
\end{equation}

At each layer $k$ of the feature extraction process, the patch size $P$ is progressively scaled, as depicted in \cref{fig:hsa}-d. This scaling adapts the sub-volume to the hierarchy of feature resolution, encoding information at multiple scales.

\myparagraph{Cross-Modal Adaptation}
To unify text and visual representations, we encourage modality interaction from early stages of the feature extraction process through two symmetric operations: Visual-to-Text Attention ($\textrm{VTA}$) and Text-to-Visual Attention ($\textrm{TVA}$).

\begin{figure}[t]
    \centering
\includegraphics[width=1.0\linewidth]{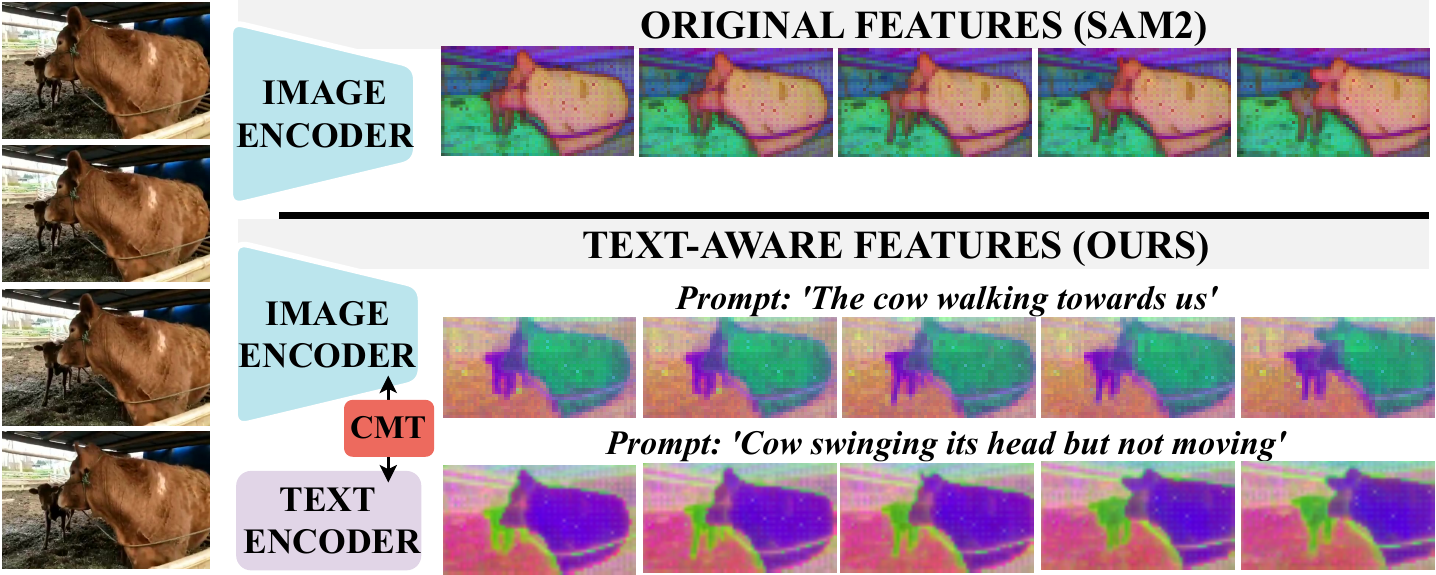}
\vspace{-0.1cm}
    \caption{\textbf{Cross Modal Temporal Adapter}: we show via PCA that our CMT provides contextualized visual features based on the given textual prompt, compared to \sam{} original ones.} 
    \label{fig:ca}
\end{figure}

Within the former, each visual feature, already enriched with temporal information through the HSA, attends to the full textual expression, allowing the model to identify candidate regions within the image based on both categorical details (\eg, the subject described in the text) and motion cues (\eg, actions), facilitating early alignment with the prompt, as visible in \cref{fig:ca}.

Formally, at layer $k$, we consider the feature of each frame in the clip, \ie $\mathcal{F}^k[t] \in \mathbb{R}^{ H_k \times W_k \times C_k}, t=1,\ldots,T$, and the set of textual embeddings $\mathcal{E}^k \in \mathbb{R}^{L \times C} $ to compute:
\begin{equation}
\mathcal{F}^{k}[t] := \mathcal{F}^{k}[t] * CA( \mathcal{F}^{k}[t], ~\mathcal{E}^k).
\end{equation}

In parallel, as the meaning of a caption can shift significantly depending on the visual content of the associated image \cite{ding_2022_vlt}, we aim at contextualizing the textual query with the semantics provided by the visual modality. To this end, the TVA progressively enriches the linguistic tokens $\mathcal{E}^k \in \mathbb{R}^{L \times C} $ with information from the visual feature maps, averaged over the video clip $\mathcal{F}^{k}_{avg}$:
\begin{equation}
\mathcal{E}^{k} := \mathcal{E}^{k} * CA( \mathcal{E}^k, ~\mathcal{F}^{k}_{avg}).
\end{equation}

\subsection{Mask prediction}
\label{sec:prediction}
At the end of the feature extraction process, we obtain the adapted visual and linguistic features, respectively $\mathcal{E}$ and $\mathcal{F}$. 
To perform the final prediction, we extract the prompt $\rho$ as in \cref{eq:prompt}, while the Memory Attention module generates the \textit{memory features} $\mathcal{F}_{mem}$ by conditioning the visual features $\mathcal{F}$ on past predictions from the Memory Bank. 
The prompt $\rho$ is fed into the frozen Mask Decoder $\mathcal{D}_{dec}$, which generates the output mask $\mathcal{P}_M \in \mathbb{R}^{1 \times H \times W}$ and the mask token $\tau_m \in \mathbb{R}^{1 \times d}$, \ie an embedding representing the segmented object. %where $d$ is decoder hidden dimensionality. 
Formally:

\begin{equation}
\begin{aligned}
    \tau_m, \mathcal{P}_M = &~\mathcal{D}_{dec} (\mathcal{F}_{mem}, \rho), \\
    \mathcal{Y} = &~\mathcal{P}_M > 0,
\end{aligned}
\end{equation}
where $\mathcal{Y} \in \mathbb{R}^{1 \times H \times W}$ denotes the output binary segmentation mask.
Finally, the Memory Encoder updates the memory bank with $\mathcal{P}_M$.

\begin{figure}[t]
    \centering
    \includegraphics[width=1.0\linewidth]{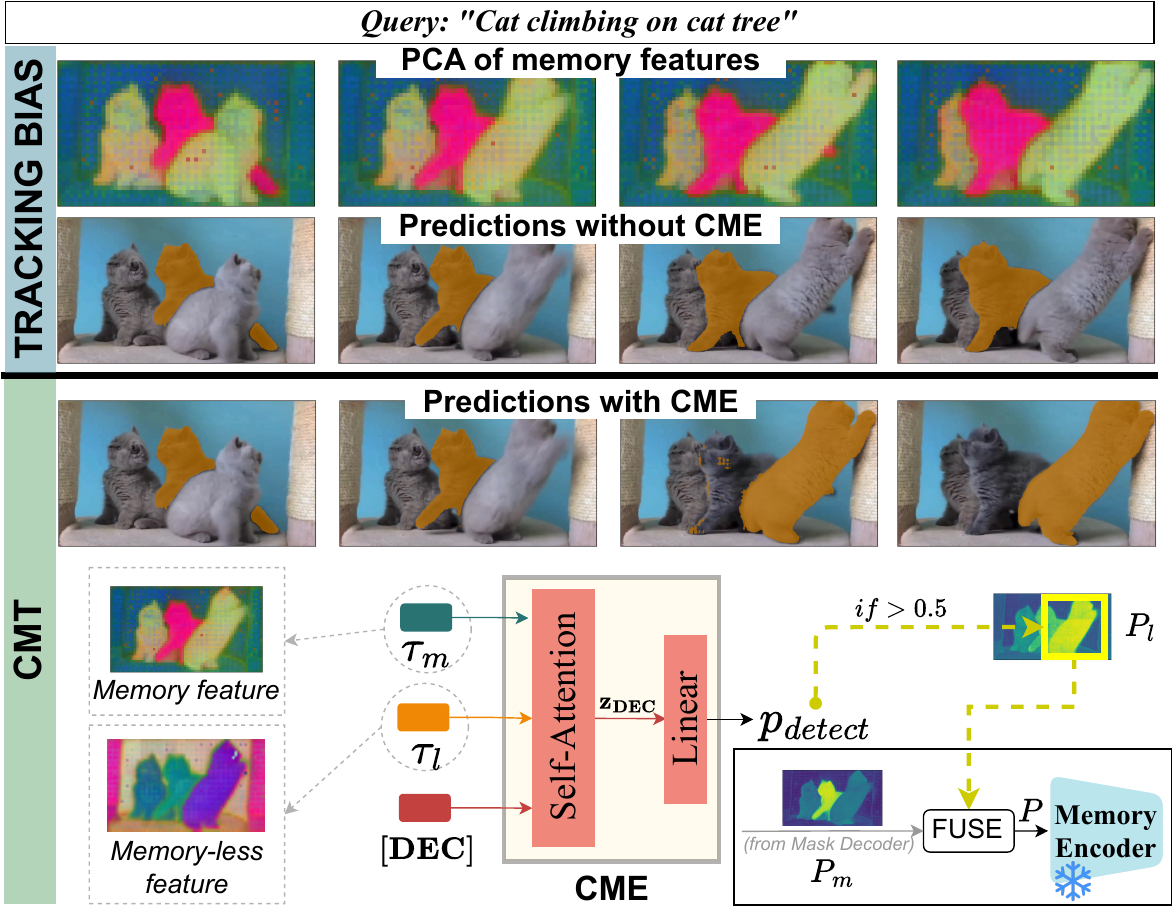}
    \vspace{-0.2cm}
    \caption{Effect of our \textbf{Conditional Memory Encoder}.
    The caption above requires disambiguating multiple instances of the same class (e.g., “cat”) by identifying a specific action (e.g., “climbing”). Since none of the instances perform this action initially, the model begins tracking the wrong instance and fails to correct itself once the target object performs the action. The top section visualizes the effect of this \textit{tracking bias} in the \textit{memory features} of \sam{}. Our CME detects that the cat that starts \textit{climbing} is more aligned with the caption, and encodes its presence in the memory bank, allowing \sam{} to switch its focus.}
    %\vspace{-0.2cm}
    \label{fig:CME}
\end{figure}

\subsection{Conditional Memory Encoder}
\label{sec:cme}
We identify as \textit{tracking bias} the phenomenon of \sam{} tracking the wrong object when the correct one is not yet identifiable in the video, and persist in following it.
This bias, as exemplified in \cref{fig:CME}, is encoded in the memory features, which are propagated to subsequent frames through the Memory Encoder.
On the other hand, we observe that the memory-less features: i) contain an \textit{unbiased} representation of the current frames, ii) are aligned with the textual prompt via our CMT (\textit{cf}. \cref{fig:ca}), and iii) can thus be used to propose candidate instances that match the prompt without being biased by past predictions. 
Building on these intuitions, we derive a \textit{memory-less token} $\tau_l$ from a cross-attention between the unbiased feature maps and the prompt. Such token represents a summary of the visual features that match the prompt. The idea is to compare it with the mask token $\tau_m$ generated by the Mask Decoder, to detect when they represent different objects, \ie, to detect when \sam{} is tracking an object that is not the one \emph{currently} most aligned with the caption. Formally:
\begin{equation}
    \tau_l = CA \left ( \mathcal{F}, \rho  \right ).
\end{equation}
We note that we initialize (and keep frozen) the weights of the cross-attention with those from \sam{} Mask Decoder.
We introduce a small learnable module, named Conditional Memory Encoder (CME), to detect such situations. When a new object is detected, a naive solution would be to compute its mask and use it to re-prompt the model at the given frame, just like a user would do, forcing \sam{} to switch its prediction. However, since the prediction computed on the \textit{memory-less} features does not have access to past video context, it might generate false positives. Thus, we propose a \textit{soft assignment}, obtained by encoding the masks of both objects in the memory bank. Essentially, the CME allows \sam{} to `see' other objects beyond the currently tracked one, and balance the influence of past context with new information, to select the one that fits the prompt the most.
In detail, our CME, illustrated in \cref{fig:CME}-bottom, concatenates the two tokens $\tau_m, \tau_l$ with a learnable \textit{decision token} $\texttt{[DEC]}$, and performs a self-attention followed by a Linear classifier:
\begin{equation}
\begin{aligned}
    \left [ z_{DEC}, z_{MT}, z_{ML} \right ]  &= SA\left( \Big [[\texttt{DEC}], \tau_m[t], \tau_l[t] \Big ] \right), \\
    p_{detect} &= \phi(z_{DEC}), \\
\end{aligned}
\end{equation}
where $\phi$ is a linear function $\mathbb{R}^{d} \rightarrow \mathbb{R}^{1} $.
When detecting a candidate text-aligned object, (\ie, $p_{detect} > 0.5$), instead of directly feeding the predicted output mask $\mathcal{P}_m$ to the Memory Encoder, our module computes the unbiased output mask, namely $\mathcal{P}_L \in \mathbb{R}^{1 \times H \times W}$, to fuse it with $\mathcal{P}_m$:
\begin{equation}
\begin{aligned}
&\mathcal{P}_l = \mathcal{D}_{dec} \left ( \mathcal{F}, \rho \right ),  \\
\mathcal{M}(h,w) &= \mathbbm{1}{(h,w)  \left [ h, w : \mathcal{P}_l > 0 \right ]},  \\
\mathcal{P} = \lambda& * \mathcal{P}_l \circ \mathcal{M} + \mathcal{P}_m \circ (1-\mathcal{M}),
\end{aligned}
\end{equation}
where $\mathcal{M}(h,w)$ is a binary mask whose value is zero except for the pixels corresponding to the object, and $\lambda$ is an hyperparameter weighing the influence of the memory-less prediction. The resulting mask $\mathcal{P}$ is fed to the Memory Encoder. 
We train the CME via self-supervision with a standard Cross-Entropy loss, by providing examples where the \textit{memory-less} features highlight different objects w.r.t the one currently tracked. We discuss in detail our training protocol in the Supp. Mat..
\begin{table*}
\begin{center}
\begin{adjustbox}{width=\linewidth}
\begin{tabular}{llllccc|ccc|cccccccccccccc}
\toprule
\multirow{2}{*}{Method} & Visual & Text & Total  & \multicolumn{3}{c|}{MeViS} & \multicolumn{3}{c|}{Ref-YouTube-VOS} & \multicolumn{3}{c}{Ref-DAVIS17} \\
\cline{5-6} \cline{7-9} \cline{10-13}
& Encoder & Encoder & Params & $\mathcal{J}$\&$\mathcal{F}$ & $\mathcal{J}$ & $\mathcal{F}$ & $\mathcal{J}$\&$\mathcal{F}$ & $\mathcal{J}$ & $\mathcal{F}$ & $\mathcal{J}$\&$\mathcal{F}$ & $\mathcal{J}$ & $\mathcal{F}$ \\
\midrule
\emph{\textbf{Large VLM} based} & & & \\
~~LISA~\cite{lai2024lisa}            \pub{CVPR'24}  & ViT-H & LLaVa  & 7 B &
37.2 & 35.1 & 39.4 & 53.9 & 53.4 & 54.3 & 64.8 & 62.2 & 67.3 \\ 
~~VISA~\cite{yan_2024_visa}           \pub{ECCV'24} & ViT-H & Chat-UniVi & 7 B &
43.5 & 40.7 & 46.3 & 61.5 & 59.8 & 63.2 & 69.4 & 66.3 & 72.5 \\ 
~~One-Token-Seg-All \cite{bai_2024_onetoken} \pub{NIPS'24} & ViT-H & Phi-3 & 3.8 B & 42.3 & 39.4 & 45.2 & 61.7 & 60.2 & 63.3 & 67.7 & 63.8 & 71.5 \\
\midrule
\midrule
~~MTTR ~\cite{botach_2022_mttr}        \pub{CVPR'22} & V-Swin-T & RoBERTa & - &
30.0 & 28.8 & 31.2 & 55.3 & 54.0 & 56.6 & - & - & -\\
~~TCE-RVOS~\cite{Hu_2024_WACV}      \pub{WACV'24}  & ResNet-50 & RoBERTa & - &
- & - & - & 59.6 & 58.3 & 60.8 & 59.4 & 56.5 & 62.4 \\
~~ReferFormer~\cite{wu_2022_language} \pub{CVPR'22} & V-Swin-B & RoBERTa & 237 M &
31.0 & 29.8 & 32.2 & 62.9 & 61.3 & 64.6 & 61.1 & 58.1 & 64.1 \\
~~SOC~\cite{luo_2024_soc}             \pub{NIPS'23}  & V-Swin-B & RoBERTa & 220 M &
- & - & - & 66.0 & 64.1 & 67.9 & 64.2 & 61.0 & 67.4 \\
~~OnlineRefer~\cite{wu_2023_online}   \pub{ICCV'23} & Swin-L   & RoBERTa & 232 M &
32.3 & 31.5 & 33.1 & 63.5 & 61.6 & 65.5 & 64.8 & 61.6 & 67.7 \\
~~LMPM~\cite{ding_2023_mevis}         \pub{ICCV'23} &  Swin-T & RoBERTa & 195 M &
37.2 & 34.2 & 40.2 & - & - & - & - & - & - \\  
~~RefSAM~\cite{li_2024_refsam} \pub{arXiv}           & ViT-B & T5 & 3 B &
- & - & - & 58.4 & 57.4 & 59.4 & 62.1 & 59.0 & 65.3 \\
~~DsHmp~\cite{he_2024_decoupling}     \pub{CVPR'24} &  V-Swin B & RoBERTa & 339 M &
- & - & - & {67.1} & {65.0} & {69.1} & {64.9} & {61.7} & {68.1}\\
~~DsHmp~\cite{he_2024_decoupling}     \pub{CVPR'24} &  Swin-T & RoBERTa & 272 M &
{46.4} & {43.0} & {49.8} & - & - & - & - & - & - \\
~~MUTR~\cite{yan2024referred}     \pub{AAAI'24} &  V-Swin-B & RoBERTa & 250 M &
- & - & - & \underline{67.5} & \underline{65.4} & \underline{69.6} & 66.4 & 62.8 & 70.0 \\

~~GroundingDINO~\cite{liu_2023_grounding}+SAM2      & Hiera-B  & BERT & 240 M &
37.7 & 34.9 & 40.5 & 57.5 & 55.6 & 59.5 & {66.4} & {62.8} & {69.9} \\
~~\ours~\textbf{(ours)}                             & Hiera-B  & CLIP-B & 150 M &
\underline{48.3} & \underline{45.4} & \underline{51.2} & {67.2} & {65.2} & {69.3} & \underline{68.5} & \underline{65.6} & \underline{71.5} \\
~~\ours~\textbf{(ours)}                             & Hiera-B  & RoBERTa & 202 M &
\textbf{49.5} & \textbf{46.6} & \textbf{52.4} & \textbf{69.2} & \textbf{67.8} & \textbf{70.6} & \textbf{70.6} & \textbf{67.4} & \textbf{74.5} \\
\bottomrule
\end{tabular}
\end{adjustbox}
\end{center}
\vspace{-0.3cm}
\caption{Comparison of \textbf{\ours{}} against state-of-the-art RVOS methods on MeViS, Ref-Youtube-VOS and Ref-DAVIS datasets. We further include methods based on large VLMs for comparison. \textbf{Bold} and \underline{underline} indicate the two top results.}
\label{tab:main_table}
\vspace{0.1cm}
\end{table*}
\section{Experimental results}
\label{sec:results}

\myparagraph{Dataset}
We evaluate our method on MeVis \cite{ding_2023_mevis}, Ref-Youtube-VOS \cite{bellver_2023_refvos} and Ref-Davis \cite{khoreva_2019_davis}.
MeViS includes 2,006 videos and features a total of 28K annotations that capture various aspects of motion. Ref-Youtube-VOS enhances the original YouTube-VOS benchmark by incorporating textual descriptions. It contains a total of 3,978 videos and approximately 15K language expressions. Ref-DAVIS17 builds upon DAVIS17 dataset, adding more than 1.5K linguistic annotations to 90 videos.

\myparagraph{Evaluation Metrics}
We utilize standard evaluation metrics, region similarity ($\mathcal{J}$), contour accuracy ($\mathcal{F}$), and their average ($\mathcal{J}$ \& $\mathcal{F}$). For MeViS and Ref-Youtube-VOS we conduct the evaluation using the official challenge servers; for Ref-DAVIS17, we used the official evaluation code.

\myparagraph{Implementation Details}
We employ Hiera-B \cite{ryali_2023_hiera} as visual extractor. As text encoder, we experiment with CLIP \cite{radford_2021_clip} and RoBERTa \cite{liu2019roberta}. We note that the text encoder and \sam{}{} weights are entirely frozen and we train only the Adapters and the CME module (4.2M parameters when using CLIP and 4.9M with RoBERTa).  Following \cite{wu_2022_language, wu_2023_online, han_2023_html, luo_2024_soc, li_2024_refsam}, we undergo pre-training for 6 epochs on RefCOCO/+/g \cite{yu_2016_refcoco, nagaraja_2016_refcocog} with a learning rate at 1e-4 and finetune on Ref-Youtube-VOS \cite{Seo_2020_urvos} for 4 epochs with a learning rate of 1e-5, using the Adam optimizer. The model trained on the Ref-YouTube-VOS is directly evaluated on DAVIS-17 \cite{khoreva_2019_davis}. On MeViS \cite{ding_2023_mevis}, we train for 1 epoch. We set $T=8$.

\subsection{Main Results}
In this section, we compare against existing works in the literature, and ablate our contributions.
In the Supp. Mat. we report additional qualitative results and ablations.

\myparagraph{Baselines}
To asses the validity of our approach, we divide the experimental comparison in the following categories:
\begin{itemize}
    \item \textbf{Standard RVOS} methods: we compare against recent relevant works in RVOS. The main comparison is \wrt the previous state-of-the-art, namely DsHmp \cite{he_2024_decoupling};
    \item Methods with \textbf{Context propagation}: OnlineRefer \cite{wu_2023_online} was the first to propose this setting. RefSAM \cite{li_2024_refsam} relies on SAM1 to provide frame-level masks, and then propagates the mask token to subsequent frames. A baseline that we propose is GroundingDINO + \sam{}, where we use the popular grounded detector to provide boxes for the first frame, and let \sam{} track the object;
    \item \textbf{Large VLM based}: Although these methods \cite{yan_2024_visa, lai2024lisa, bai_2024_onetoken} are not comparable to ours, or previous ones, in terms of model size, we include them in the table to provide an interesting reference of performance.
\end{itemize}   
%These results are reported in \cref{tab:main_table}, and discussed below.

\myparagraph{Comparison with standard RVOS methods}
Traditional RVOS methods, such as ReferFormer \cite{wu_2022_language} and MTTR \cite{botach_2022_mttr}, suffer a significant performance drop on the MeViS benchmark, as they are unable to solve queries which require to model long-term context.
%process clips independently, without keeping memory of the global video context. As argued in recent works \cite{ding_2023_mevis, he_2024_decoupling}, this characteristic leads to a significant performance drop on the MeViS benchmark. 
An exception is represented by LMPM and its follow up work DsHmp, which represents the state-of-the-art: these methods process the entire video clip at once, modeling multiple trajectories for all the instances in the video to select the one that fits the prompt the most. %While this boosts their performance, it also means they cannot be deployed in applications that require streaming processing.
Despite this, \ours{} outperforms DsHmp \cite{he_2024_decoupling} on all three datasets, improving $\mathcal{J}\&\mathcal{F}$  of +3.1\%, +2.1\%, and +5.7\%, while utilizing a smaller model in terms of total parameters. Notably, we achieve this by training only 4.9 M parameters out of 202 M. This result is particularly impressive, as offline methods exploit information from the entire video to handle challenges such as late-appearing objects or motion-dependent disambiguation, as opposed to our streaming approach. 
With respect to other methods, we outperform them significantly on MeViS, whereas the gap is smaller on Ref-YouTube-VOS and Ref-DAVIS, which contain more descriptive captions, and object-centric videos.
Lastly, we also experiment with the text encoder of CLIP, which achieves state-of-the-art results on MeViS, and competitive performance on other benchmarks, while providing a more compact model with just 150 M params.

\myparagraph{Comparison with Context Propagation methods}
%MeViS presents the greatest challenge as it emphasizes motion-specific expressions to disambiguate between similar-looking instances \cite{ding_2023_mevis}.
Our proposed baseline GroundingDINO \cite{liu_2023_grounding}+SAM2, while obviously flawed, being forced to predict the desired instance based on the first frame only, achieves acceptable results on Ref-DAVIS, whereas on MeViS and Ref-Youtube-VOS its performance drops of 11.8\% and 11.7\%, respectively.
Differently, \ours{}, demonstrates  excellent performance in both motion-dependent and static scenarios. Specifically, on MeViS, we outperform OnlineRefer \cite{wu_2023_online} by +17.2\%. On Ref-YouTube-VOS and Ref-DAVIS, the gap is of +5.7\%, and +5.8\%, respectively. 

\myparagraph{Comparison with Large-VLM based} 
While comparisons with Large-VLM based approaches are not standard in RVOS evaluations, we include them in this work to provide additional context. The VLM-based solutions \cite{lai2024lisa, yan_2024_visa, bai_2024_onetoken} are designed to leverage the extensive reasoning capabilities of VLMs to address complex textual instructions and implicit descriptions that require world knowledge \cite{lai2024lisa}. This leads to improved performance in tasks like MeViS, where reasoning over motion patterns is required. However, 
delegating cross-modal reasoning to these VLMs incurs in significant computational overhead, whereas \ours{} incorporates visual-text interaction directly at the feature level.
Notably, \ours{} outperforms VISA, the best VLM-based competitor by a substantial margin, respectively +6\%, +7.7\%, +2.9\% on the three benchmarks.

\begin{table}
\begin{center}
\begin{adjustbox}{width=0.95\linewidth}
% \centering
\setlength{\tabcolsep}{4pt}
\begin{tabular}{ccccc|c}
\toprule
\rowcolor{cyan!10} MLP-only  & Text-to-Visual & Visual-to-Text & HSA & CME & $\mathcal{J}$\&$\mathcal{F}$ \\
\midrule
\checkmark &            &            &         &   & 45.2 \\
\checkmark & \checkmark &            &          &  & 47.5 \\
\checkmark &            & \checkmark &         &   & 48.3 \\
\checkmark & \checkmark & \checkmark &         &   & 50.3 \\
\checkmark & \checkmark & \checkmark & \checkmark & & 54.2 \\
\midrule
\checkmark & \checkmark & \checkmark & \checkmark & \checkmark  & 55.5 \\
\bottomrule
\end{tabular}
\end{adjustbox}
\end{center}
\vspace{-0.3cm}
\caption{Ablation of our \textbf{Cross-Modal Temporal Adapter} (CMT). We show the effect of not using CMT (\ie \textit{MLP-only} to prompt \sam{}), vs. adding one at a time its core components. Lastly, the \textbf{Conditional Memory Encoder} (CME) is added.}
%\vspace{-0.2cm}
\label{tab:ablate_adapt}
\end{table}

\subsection{Ablation Studies}
We conduct our ablations on MeViS, as it embodies the core challenges of \textit{online} RVOS.
We report results on the `valid\_u' set \cite{ding_2023_mevis}, employing CLIP-B as text encoder.

\myparagraph{Making SAM2 Wiser}
We start by showing, in \cref{tab:ablate_adapt}, how each of the core components of our CMT Adapter progressively injects \textit{wisdom} (\ie knowledge about language and temporal context) into SAM2.
The first line reports the result using the `naive' solution of aligning the textual prompt to the visual features using a single learnable MLP \cite{zhu_2023_minigpt}. While effective to some extent, the results show that allowing early interaction of the two modalities grants a substantial boost (+5.1\% with both adapters). Adding explicit temporal feature modeling provides an additional improvements of +3.9\%.
These results sustain our intuition that adding frame-level alignment through a MLP is not enough to obtain robust performances, and that it is essential to couple the text and visual semantics, as well as modeling temporal context.
Lastly, the table shows how adding our CME is effective in mitigating SAM2 \textit{tracking bias}. 

\begin{table}
\begin{center}
\begin{adjustbox}{width=.75\linewidth}
\begin{tabular}{c|cccccc}
\toprule
\rowcolor{cyan!10} & \multicolumn{5}{c}{HSA Patch Size} \\
& \multicolumn{3}{c|}{{\begin{tabular}[c]{@{}c@{}}Fixed Size\end{tabular}}} & \multicolumn{2}{c}{{\begin{tabular}[c]{@{}c@{}}Hierarchical\end{tabular}}} \\
\cline{2-6}
 & 1 & 4 & \multicolumn{1}{c|}{8} & 8 / 4 / 2 & 16 / 8 / 4 \\
\midrule
$\mathcal{J}$\&$\mathcal{F}$ & 49.7 & 52.3 & \multicolumn{1}{c|}{53.1} & \textbf{54.2} & \underline{53.8}  \\
\midrule
\midrule
\rowcolor{cyan!10} & \multicolumn{5}{c}{CME vs Random choice} \\
 & Never & \multicolumn{2}{c}{Always}  & 1 every 4 & CME  \\
\midrule
$\mathcal{J}$\&$\mathcal{F}$ & 54.2 & \multicolumn{2}{c}{50.7} & 52.4 & \textbf{55.5}   \\
\bottomrule
\end{tabular}
\end{adjustbox}
\end{center}
\vspace{-0.3cm}
\caption{Top: Ablation of the Patch size in our \textbf{HSA}, with the effect of a fixed size vs Hierarchical. Bottom: Effect of not predicting detections (\textit{Never}), predicting them at every frame (\textit{Always}), randomly (\textit{1 in 4}) vs. using predictions of our CME.}
\label{tab:hsa_cme}
%\vspace{-0.5cm}
\end{table}
\myparagraph{Hierarchical Selective Attention}
In the top section of \cref{tab:hsa_cme}, we study the effect of the spatial patch size in our HSA module, which models the temporal evolution of over a spatial patch of size $P$ across the temporal axis.
Using $P=1$ is equivalent to processing each pixel independently across frames. The table shows that including spatial context, up to 8 pixels, is beneficial.
Using a hierarchical patch size that scales with the feature map resolution yields a gain of +1.1\% over the fixed sized alternative.

\myparagraph{Conditional Memory Encoder}
The bottom section of \cref{tab:hsa_cme} provides insight into our CME module.
The CME, essentially, detects whenever an object in the \textit{unbiased} feature maps of the current frame displays higher alignment with the textual prompts \wrt the currently tracked one, but SAM2 fails in noticing it due to the \textit{tracking bias} (\cref{fig:CME}). 
The table compares the effect of \textit{Never} applying such strategy (\ie, not using CME), doing it \textit{Always} (\ie, at every frame), or once every 4 frames.
The results show that increasing the frequency of \textit{artificial} detection worsens performances, adding noise to the tracking, whereas the predictions of our CME are beneficial, with a boost of +1.3\%.

\myparagraph{Adaptation strategy}
In \cref{tab:adapters} we compare different adaptation strategies, including Full Fine-Tuning (FT), LoRa \cite{hu2022lora}, AdaptFormer \cite{chen2022adaptformer}, and the proposed CMT.

\begin{table}[h]
\setlength{\tabcolsep}{10pt}
\begin{center}
\begin{adjustbox}{width=0.95\linewidth}
% \centering
\setlength{\tabcolsep}{2pt}
\begin{tabular}{lcccccccc}
\toprule
\rowcolor{cyan!10} &  Full-FT  &  LoRa \cite{hu2022lora} &  AdaptFormer \cite{chen2022adaptformer} &  CMT (ours) \\
\midrule
MeViS $\mathcal{J}$\&$\mathcal{F}$   & 43.1 & 44.2 & 43.9  & \textbf{48.3} \\
\bottomrule
\end{tabular}
\end{adjustbox}
\end{center}
\vspace{-0.3cm}
\caption{Baselines: prior adapters and full-finetune, with CLIP-B.}
\label{tab:adapters}
\vspace{-0.36cm}
\end{table}

\section{Conclusion}
\label{sec:conclusion}

In this work, we introduced \ours, a novel approach for RVOS that builds upon the SAM2 model by incorporating i) natural language understanding, ii) temporal feature modeling, and iii) a learnable strategy to adjusts tracking focus according to visual cues that emerge over time.
\ours{} achieves SOTA performance across benchmarks while adding less than 5M parameters, without modifying SAM2 weights or using external models for visual-text alignment.
We obtain an effective pipeline for applications of streaming video segmentation, addressing limitations of existing RVOS approaches, which either lack long-term context or rely on single-frame context propagation.

\small{
\myparagraph{Acknowledgements}
Claudia Cuttano was supported by the Sustainable Mobility Center (CNMS) which received funding from the European Union Next Generation EU (Piano Nazionale di Ripresa e Resilienza (PNRR), Missione 4 Componente 2 Investimento 1.4 "Potenziamento strutture di ricerca e creazione di "campioni nazionali di R\&S" su alcune Key Enabling Technologies") with grant agreement no. CN\_00000023. 

\noindent Gabriele Trivigno, Giuseppe Averta and Carlo Masone were supported by FAIR - Future Artificial Intelligence Research which received funding from the European Union Next-GenerationEU (PIANO NAZIONALE DI RIPRESA E RESILIENZA (PNRR) – MISSIONE 4 COMPONENTE 2, INVESTIMENTO 1.3 – D.D. 1555 11/10/2022, PE00000013).  This manuscript reflects only the authors’ views and opinions, neither the European Union nor the European Commission can be considered responsible for them.

\noindent We acknowledge the CINECA award
under the ISCRA initiative, for the availability of high per-
formance computing resources.}

{
    \small
    \bibliographystyle{ieeenat_fullname}
    \bibliography{main}

\begin{thebibliography}{52}
\providecommand{\natexlab}[1]{#1}
\providecommand{\url}[1]{\texttt{#1}}
\expandafter\ifx\csname urlstyle\endcsname\relax
  \providecommand{\doi}[1]{doi: #1}\else
  \providecommand{\doi}{doi: \begingroup \urlstyle{rm}\Url}\fi

\bibitem[Bai et~al.(2024)Bai, He, Mei, Wang, Gao, Chen, Liu, Zhang, and Shou]{bai_2024_onetoken}
Zechen Bai, Tong He, Haiyang Mei, Pichao Wang, Ziteng Gao, Joya Chen, Lei Liu, Zheng Zhang, and Mike~Zheng Shou.
\newblock One token to seg them all: Language instructed reasoning segmentation in videos.
\newblock \emph{arXiv preprint arXiv:2409.19603}, 2024.

\bibitem[Bellver et~al.(2023)Bellver, Ventura, Silberer, Kazakos, Torres, and Giro-i Nieto]{bellver_2023_refvos}
Miriam Bellver, Carles Ventura, Carina Silberer, Ioannis Kazakos, Jordi Torres, and Xavier Giro-i Nieto.
\newblock A closer look at referring expressions for video object segmentation.
\newblock \emph{Multimedia Tools and Applications}, 82\penalty0 (3):\penalty0 4419--4438, 2023.

\bibitem[Botach et~al.(2022)Botach, Zheltonozhskii, and Baskin]{botach_2022_mttr}
Adam Botach, Evgenii Zheltonozhskii, and Chaim Baskin.
\newblock End-to-end referring video object segmentation with multimodal transformers.
\newblock In \emph{Proceedings of the IEEE/CVF Conference on Computer Vision and Pattern Recognition}, pages 4985--4995, 2022.

\bibitem[Carion et~al.(2020)Carion, Massa, Synnaeve, Usunier, Kirillov, and Zagoruyko]{carion_2020_end}
Nicolas Carion, Francisco Massa, Gabriel Synnaeve, Nicolas Usunier, Alexander Kirillov, and Sergey Zagoruyko.
\newblock End-to-end object detection with transformers.
\newblock In \emph{European conference on computer vision}, pages 213--229. Springer, 2020.

\bibitem[Chen et~al.(2022)Chen, Ge, Tong, Wang, Song, Wang, and Luo]{chen2022adaptformer}
Shoufa Chen, Chongjian Ge, Zhan Tong, Jiangliu Wang, Yibing Song, Jue Wang, and Ping Luo.
\newblock Adaptformer: Adapting vision transformers for scalable visual recognition.
\newblock \emph{Advances in Neural Information Processing Systems}, 35:\penalty0 16664--16678, 2022.

\bibitem[Chng et~al.(2024)Chng, Zheng, Han, Qiu, and Huang]{chng2024mask}
Yong~Xien Chng, Henry Zheng, Yizeng Han, Xuchong Qiu, and Gao Huang.
\newblock Mask grounding for referring image segmentation.
\newblock In \emph{Proceedings of the IEEE/CVF Conference on Computer Vision and Pattern Recognition}, pages 26573--26583, 2024.

\bibitem[Ding et~al.(2022)Ding, Liu, Wang, and Jiang]{ding_2022_vlt}
Henghui Ding, Chang Liu, Suchen Wang, and Xudong Jiang.
\newblock Vlt: Vision-language transformer and query generation for referring segmentation.
\newblock \emph{IEEE Transactions on Pattern Analysis and Machine Intelligence}, 45\penalty0 (6):\penalty0 7900--7916, 2022.

\bibitem[Ding et~al.(2023)Ding, Liu, He, Jiang, and Loy]{ding_2023_mevis}
Henghui Ding, Chang Liu, Shuting He, Xudong Jiang, and Chen~Change Loy.
\newblock {MeViS}: A large-scale benchmark for video segmentation with motion expressions.
\newblock In \emph{ICCV}, 2023.

\bibitem[Ding et~al.(2021)Ding, Hui, Huang, Liu, Luo, Huang, and Wei]{ding_2021_pminet}
Zihan Ding, Tianrui Hui, Shaofei Huang, Si Liu, Xuan Luo, Junshi Huang, and Xiaoming Wei.
\newblock Progressive multimodal interaction network for referring video object segmentation.
\newblock \emph{The 3rd Large-scale Video Object Segmentation Challenge}, 8\penalty0 (10), 2021.

\bibitem[Gavrilyuk et~al.(2018)Gavrilyuk, Ghodrati, Li, and Snoek]{gavrilyuk_2018_actor}
Kirill Gavrilyuk, Amir Ghodrati, Zhenyang Li, and Cees~GM Snoek.
\newblock Actor and action video segmentation from a sentence.
\newblock In \emph{Proceedings of the IEEE conference on computer vision and pattern recognition}, pages 5958--5966, 2018.

\bibitem[Han et~al.(2023)Han, Wang, Li, Yao, Chang, and Qiao]{han_2023_html}
Mingfei Han, Yali Wang, Zhihui Li, Lina Yao, Xiaojun Chang, and Yu Qiao.
\newblock Html: Hybrid temporal-scale multimodal learning framework for referring video object segmentation.
\newblock In \emph{Proceedings of the IEEE/CVF International Conference on Computer Vision}, pages 13414--13423, 2023.

\bibitem[He and Ding(2024)]{he_2024_decoupling}
Shuting He and Henghui Ding.
\newblock Decoupling static and hierarchical motion perception for referring video segmentation.
\newblock In \emph{Proceedings of the IEEE/CVF Conference on Computer Vision and Pattern Recognition}, pages 13332--13341, 2024.

\bibitem[Houlsby et~al.(2019)Houlsby, Giurgiu, Jastrzebski, Morrone, De~Laroussilhe, Gesmundo, Attariyan, and Gelly]{houlsby_2019_adapter}
Neil Houlsby, Andrei Giurgiu, Stanislaw Jastrzebski, Bruna Morrone, Quentin De~Laroussilhe, Andrea Gesmundo, Mona Attariyan, and Sylvain Gelly.
\newblock Parameter-efficient transfer learning for nlp.
\newblock In \emph{International conference on machine learning}, pages 2790--2799. PMLR, 2019.

\bibitem[Hu et~al.(2022)Hu, Shen, Wallis, Allen-Zhu, Li, Wang, Wang, Chen, et~al.]{hu2022lora}
Edward~J Hu, Yelong Shen, Phillip Wallis, Zeyuan Allen-Zhu, Yuanzhi Li, Shean Wang, Lu Wang, Weizhu Chen, et~al.
\newblock Lora: Low-rank adaptation of large language models.
\newblock \emph{ICLR}, 1\penalty0 (2):\penalty0 3, 2022.

\bibitem[Hu et~al.(2024)Hu, Hampiholi, Neumann, and Lang]{Hu_2024_WACV}
Xiao Hu, Basavaraj Hampiholi, Heiko Neumann, and Jochen Lang.
\newblock Temporal context enhanced referring video object segmentation.
\newblock In \emph{Proceedings of the IEEE/CVF Winter Conference on Applications of Computer Vision (WACV)}, pages 5574--5583, 2024.

\bibitem[Jiang et~al.(2022)Jiang, Zhang, Huang, Ge, Ni, Lu, Zhou, Song, and Huang]{jiang_2022_cross}
Haojun Jiang, Jianke Zhang, Rui Huang, Chunjiang Ge, Zanlin Ni, Jiwen Lu, Jie Zhou, Shiji Song, and Gao Huang.
\newblock Cross-modal adapter for text-video retrieval.
\newblock \emph{arXiv preprint arXiv:2211.09623}, 2022.

\bibitem[Jin et~al.(2024)Jin, Zhang, Gong, Xu, Deng, Wang, Zhang, Shen, and Feng]{jin_2024_mv}
Xiaojie Jin, Bowen Zhang, Weibo Gong, Kai Xu, Xueqing Deng, Peng Wang, Zhao Zhang, Xiaohui Shen, and Jiashi Feng.
\newblock Mv-adapter: Multimodal video transfer learning for video text retrieval.
\newblock In \emph{Proceedings of the IEEE/CVF Conference on Computer Vision and Pattern Recognition}, pages 27144--27153, 2024.

\bibitem[Khoreva et~al.(2019)Khoreva, Rohrbach, and Schiele]{khoreva_2019_davis}
Anna Khoreva, Anna Rohrbach, and Bernt Schiele.
\newblock Video object segmentation with language referring expressions.
\newblock In \emph{Computer Vision--ACCV 2018: 14th Asian Conference on Computer Vision, Perth, Australia, December 2--6, 2018, Revised Selected Papers, Part IV 14}, pages 123--141. Springer, 2019.

\bibitem[Kirillov et~al.(2023)Kirillov, Mintun, Ravi, Mao, Rolland, Gustafson, Xiao, Whitehead, Berg, Lo, et~al.]{kirillov_2023_sam}
Alexander Kirillov, Eric Mintun, Nikhila Ravi, Hanzi Mao, Chloe Rolland, Laura Gustafson, Tete Xiao, Spencer Whitehead, Alexander~C Berg, Wan-Yen Lo, et~al.
\newblock Segment anything.
\newblock In \emph{Proceedings of the IEEE/CVF International Conference on Computer Vision}, pages 4015--4026, 2023.

\bibitem[Lai et~al.(2024)Lai, Tian, Chen, Li, Yuan, Liu, and Jia]{lai2024lisa}
Xin Lai, Zhuotao Tian, Yukang Chen, Yanwei Li, Yuhui Yuan, Shu Liu, and Jiaya Jia.
\newblock Lisa: Reasoning segmentation via large language model.
\newblock In \emph{Proceedings of the IEEE/CVF Conference on Computer Vision and Pattern Recognition}, pages 9579--9589, 2024.

\bibitem[Li et~al.(2024)Li, Zhang, Teng, Lan, and Liu]{li_2024_refsam}
Yonglin Li, Jing Zhang, Xiao Teng, Long Lan, and Xinwang Liu.
\newblock Refsam: Efficiently adapting segmenting anything model for referring video object segmentation, 2024.

\bibitem[Liu et~al.(2024{\natexlab{a}})Liu, Li, Wu, and Lee]{liu_2024_llava}
Haotian Liu, Chunyuan Li, Qingyang Wu, and Yong~Jae Lee.
\newblock Visual instruction tuning.
\newblock \emph{Advances in neural information processing systems}, 36, 2024{\natexlab{a}}.

\bibitem[Liu et~al.(2023{\natexlab{a}})Liu, Huang, Li, Feng, Wu, and Li]{liu_2023_revisiting}
Ruyang Liu, Jingjia Huang, Ge Li, Jiashi Feng, Xinglong Wu, and Thomas~H Li.
\newblock Revisiting temporal modeling for clip-based image-to-video knowledge transferring.
\newblock In \emph{Proceedings of the IEEE/CVF Conference on Computer Vision and Pattern Recognition}, pages 6555--6564, 2023{\natexlab{a}}.

\bibitem[Liu et~al.(2024{\natexlab{b}})Liu, Li, Ge, Li, Shan, and Li]{Liu_2024_btadapter}
Ruyang Liu, Chen Li, Yixiao Ge, Thomas~H. Li, Ying Shan, and Ge Li.
\newblock Bt-adapter: Video conversation is feasible without video instruction tuning.
\newblock In \emph{Proceedings of the IEEE/CVF Conference on Computer Vision and Pattern Recognition (CVPR)}, pages 13658--13667, 2024{\natexlab{b}}.

\bibitem[Liu et~al.(2021)Liu, Hui, Huang, Wei, Li, and Li]{liu_2021_cmpc}
Si Liu, Tianrui Hui, Shaofei Huang, Yunchao Wei, Bo Li, and Guanbin Li.
\newblock Cross-modal progressive comprehension for referring segmentation.
\newblock \emph{IEEE Transactions on Pattern Analysis and Machine Intelligence}, 44\penalty0 (9):\penalty0 4761--4775, 2021.

\bibitem[Liu et~al.(2023{\natexlab{b}})Liu, Zeng, Ren, Li, Zhang, Yang, Li, Yang, Su, Zhu, et~al.]{liu_2023_grounding}
Shilong Liu, Zhaoyang Zeng, Tianhe Ren, Feng Li, Hao Zhang, Jie Yang, Chunyuan Li, Jianwei Yang, Hang Su, Jun Zhu, et~al.
\newblock Grounding dino: Marrying dino with grounded pre-training for open-set object detection.
\newblock \emph{arXiv preprint arXiv:2303.05499}, 2023{\natexlab{b}}.

\bibitem[Liu et~al.(2019)Liu, Ott, Goyal, Du, Joshi, Chen, Levy, Lewis, Zettlemoyer, and Stoyanov]{liu2019roberta}
Yinhan Liu, Myle Ott, Naman Goyal, Jingfei Du, Mandar Joshi, Danqi Chen, Omer Levy, Mike Lewis, Luke Zettlemoyer, and Veselin Stoyanov.
\newblock Roberta: A robustly optimized bert pretraining approach.
\newblock \emph{arXiv preprint arXiv:1907.11692}, 2019.

\bibitem[Lu et~al.(2023)Lu, Huo, Yang, Lu, Zhan, Tomizuka, and Ding]{lu_2023_uniadapter}
Haoyu Lu, Yuqi Huo, Guoxing Yang, Zhiwu Lu, Wei Zhan, Masayoshi Tomizuka, and Mingyu Ding.
\newblock Uniadapter: Unified parameter-efficient transfer learning for cross-modal modeling.
\newblock \emph{arXiv preprint arXiv:2302.06605}, 2023.

\bibitem[Luo et~al.(2024)Luo, Xiao, Liu, Li, Wang, Tang, Li, and Yang]{luo_2024_soc}
Zhuoyan Luo, Yicheng Xiao, Yong Liu, Shuyan Li, Yitong Wang, Yansong Tang, Xiu Li, and Yujiu Yang.
\newblock Soc: Semantic-assisted object cluster for referring video object segmentation.
\newblock \emph{Advances in Neural Information Processing Systems}, 36, 2024.

\bibitem[Miao et~al.(2023)Miao, Bennamoun, Gao, and Mian]{miao_2023_sgmg}
Bo Miao, Mohammed Bennamoun, Yongsheng Gao, and Ajmal Mian.
\newblock Spectrum-guided multi-granularity referring video object segmentation.
\newblock In \emph{Proceedings of the IEEE/CVF International Conference on Computer Vision}, pages 920--930, 2023.

\bibitem[Nagaraja et~al.(2016)Nagaraja, Morariu, and Davis]{nagaraja_2016_refcocog}
Varun~K Nagaraja, Vlad~I Morariu, and Larry~S Davis.
\newblock Modeling context between objects for referring expression understanding.
\newblock In \emph{Computer Vision--ECCV 2016: 14th European Conference, Amsterdam, The Netherlands, October 11--14, 2016, Proceedings, Part IV 14}, pages 792--807. Springer, 2016.

\bibitem[Oh et~al.(2019)Oh, Lee, Xu, and Kim]{oh_2019_memory}
Seoung~Wug Oh, Joon-Young Lee, Ning Xu, and Seon~Joo Kim.
\newblock Video object segmentation using space-time memory networks.
\newblock In \emph{Proceedings of the IEEE/CVF international conference on computer vision}, pages 9226--9235, 2019.

\bibitem[Patrick et~al.(2021)Patrick, Campbell, Asano, Misra, Metze, Feichtenhofer, Vedaldi, and Henriques]{patrick_2021_keeping}
Mandela Patrick, Dylan Campbell, Yuki Asano, Ishan Misra, Florian Metze, Christoph Feichtenhofer, Andrea Vedaldi, and Joao~F Henriques.
\newblock Keeping your eye on the ball: Trajectory attention in video transformers.
\newblock \emph{Advances in neural information processing systems}, 34:\penalty0 12493--12506, 2021.

\bibitem[Peters et~al.(2019)Peters, Ruder, and Smith]{peters_2019_tune}
Matthew~E Peters, Sebastian Ruder, and Noah~A Smith.
\newblock To tune or not to tune? adapting pretrained representations to diverse tasks.
\newblock \emph{arXiv preprint arXiv:1903.05987}, 2019.

\bibitem[Radford et~al.(2021)Radford, Kim, Hallacy, Ramesh, Goh, Agarwal, Sastry, Askell, Mishkin, Clark, et~al.]{radford_2021_clip}
Alec Radford, Jong~Wook Kim, Chris Hallacy, Aditya Ramesh, Gabriel Goh, Sandhini Agarwal, Girish Sastry, Amanda Askell, Pamela Mishkin, Jack Clark, et~al.
\newblock Learning transferable visual models from natural language supervision.
\newblock In \emph{International conference on machine learning}, pages 8748--8763. PMLR, 2021.

\bibitem[Ravi et~al.(2024)Ravi, Gabeur, Hu, Hu, Ryali, Ma, Khedr, R{\"a}dle, Rolland, Gustafson, et~al.]{ravi_2024_sam2}
Nikhila Ravi, Valentin Gabeur, Yuan-Ting Hu, Ronghang Hu, Chaitanya Ryali, Tengyu Ma, Haitham Khedr, Roman R{\"a}dle, Chloe Rolland, Laura Gustafson, et~al.
\newblock Sam 2: Segment anything in images and videos.
\newblock \emph{arXiv preprint arXiv:2408.00714}, 2024.

\bibitem[Ren et~al.(2024)Ren, Liu, Zeng, Lin, Li, Cao, Chen, Huang, Chen, Yan, et~al.]{ren_2024_groundedsam}
Tianhe Ren, Shilong Liu, Ailing Zeng, Jing Lin, Kunchang Li, He Cao, Jiayu Chen, Xinyu Huang, Yukang Chen, Feng Yan, et~al.
\newblock Grounded sam: Assembling open-world models for diverse visual tasks.
\newblock \emph{arXiv preprint arXiv:2401.14159}, 2024.

\bibitem[Ryali et~al.(2023)Ryali, Hu, Bolya, Wei, Fan, Huang, Aggarwal, Chowdhury, Poursaeed, Hoffman, et~al.]{ryali_2023_hiera}
Chaitanya Ryali, Yuan-Ting Hu, Daniel Bolya, Chen Wei, Haoqi Fan, Po-Yao Huang, Vaibhav Aggarwal, Arkabandhu Chowdhury, Omid Poursaeed, Judy Hoffman, et~al.
\newblock Hiera: A hierarchical vision transformer without the bells-and-whistles.
\newblock In \emph{International Conference on Machine Learning}, pages 29441--29454. PMLR, 2023.

\bibitem[Seo et~al.(2020)Seo, Lee, and Han]{Seo_2020_urvos}
Seonguk Seo, Joon-Young Lee, and Bohyung Han.
\newblock Urvos: Unified referring video object segmentation network with a large-scale benchmark.
\newblock In \emph{European Conference on Computer Vision}, 2020.

\bibitem[Tang et~al.(2023)Tang, Zheng, and Yang]{tang_2023_tempcd}
Jiajin Tang, Ge Zheng, and Sibei Yang.
\newblock Temporal collection and distribution for referring video object segmentation.
\newblock In \emph{Proceedings of the IEEE/CVF International Conference on Computer Vision}, pages 15466--15476, 2023.

\bibitem[Wang et~al.(2024)Wang, Xing, Jiang, Chen, Mei, Zuo, Dai, Wang, and Liu]{wang_2024_multimodal}
Mengmeng Wang, Jiazheng Xing, Boyuan Jiang, Jun Chen, Jianbiao Mei, Xingxing Zuo, Guang Dai, Jingdong Wang, and Yong Liu.
\newblock A multimodal, multi-task adapting framework for video action recognition.
\newblock In \emph{Proceedings of the AAAI Conference on Artificial Intelligence}, pages 5517--5525, 2024.

\bibitem[Wang et~al.(2022)Wang, Bao, Dong, Bjorck, Peng, Liu, Aggarwal, Mohammed, Singhal, Som, et~al.]{wang_2022_beit}
Wenhui Wang, Hangbo Bao, Li Dong, Johan Bjorck, Zhiliang Peng, Qiang Liu, Kriti Aggarwal, Owais~Khan Mohammed, Saksham Singhal, Subhojit Som, et~al.
\newblock Image as a foreign language: Beit pretraining for all vision and vision-language tasks.
\newblock \emph{arXiv preprint arXiv:2208.10442}, 2022.

\bibitem[Wu et~al.(2023)Wu, Wang, Zhang, Zhang, and Shen]{wu_2023_online}
Dongming Wu, Tiancai Wang, Yuang Zhang, Xiangyu Zhang, and Jianbing Shen.
\newblock {OnlineRefer}: A simple online baseline for referring video object segmentation.
\newblock In \emph{Proceedings of the IEEE/CVF International Conference on Computer Vision}, pages 2761--2770, 2023.

\bibitem[Wu et~al.(2022)Wu, Jiang, Sun, Yuan, and Luo]{wu_2022_language}
Jiannan Wu, Yi Jiang, Peize Sun, Zehuan Yuan, and Ping Luo.
\newblock Language as queries for referring video object segmentation.
\newblock In \emph{Proceedings of the IEEE/CVF Conference on Computer Vision and Pattern Recognition}, pages 4974--4984, 2022.

\bibitem[Xu et~al.(2023)Xu, Chen, Zhang, Song, Wan, and Li]{xu_2023_bridging}
Zunnan Xu, Zhihong Chen, Yong Zhang, Yibing Song, Xiang Wan, and Guanbin Li.
\newblock Bridging vision and language encoders: Parameter-efficient tuning for referring image segmentation.
\newblock In \emph{Proceedings of the IEEE/CVF International Conference on Computer Vision}, pages 17503--17512, 2023.

\bibitem[Yan et~al.(2024{\natexlab{a}})Yan, Wang, Yan, Jiang, Hu, Kang, Xie, and Gavves]{yan_2024_visa}
Cilin Yan, Haochen Wang, Shilin Yan, Xiaolong Jiang, Yao Hu, Guoliang Kang, Weidi Xie, and Efstratios Gavves.
\newblock Visa: Reasoning video object segmentation via large language models.
\newblock \emph{arXiv preprint arXiv:2407.11325}, 2024{\natexlab{a}}.

\bibitem[Yan et~al.(2024{\natexlab{b}})Yan, Zhang, Guo, Chen, Zhang, Li, Qiao, Dong, He, and Gao]{yan2024referred}
Shilin Yan, Renrui Zhang, Ziyu Guo, Wenchao Chen, Wei Zhang, Hongyang Li, Yu Qiao, Hao Dong, Zhongjiang He, and Peng Gao.
\newblock Referred by multi-modality: A unified temporal transformer for video object segmentation.
\newblock In \emph{Proceedings of the AAAI Conference on Artificial Intelligence}, pages 6449--6457, 2024{\natexlab{b}}.

\bibitem[Yang et~al.(2024)Yang, Zhang, Wang, and Xie]{yang_2024_mma}
Lingxiao Yang, Ru-Yuan Zhang, Yanchen Wang, and Xiaohua Xie.
\newblock Mma: Multi-modal adapter for vision-language models.
\newblock In \emph{Proceedings of the IEEE/CVF Conference on Computer Vision and Pattern Recognition}, pages 23826--23837, 2024.

\bibitem[Ye et~al.(2019)Ye, Rochan, Liu, and Wang]{ye_2019_cmsa}
Linwei Ye, Mrigank Rochan, Zhi Liu, and Yang Wang.
\newblock Cross-modal self-attention network for referring image segmentation.
\newblock In \emph{Proceedings of the IEEE/CVF conference on computer vision and pattern recognition}, pages 10502--10511, 2019.

\bibitem[Yu et~al.(2016)Yu, Poirson, Yang, Berg, and Berg]{yu_2016_refcoco}
Licheng Yu, Patrick Poirson, Shan Yang, Alexander~C Berg, and Tamara~L Berg.
\newblock Modeling context in referring expressions.
\newblock In \emph{Computer Vision--ECCV 2016: 14th European Conference, Amsterdam, The Netherlands, October 11-14, 2016, Proceedings, Part II 14}, pages 69--85. Springer, 2016.

\bibitem[Zhang et~al.(2024)Zhang, Cheng, Hu, Liu, Ran, Chen, Liu, Wang, et~al.]{zhang_2024_evf}
Yuxuan Zhang, Tianheng Cheng, Rui Hu, Heng Liu, Longjin Ran, Xiaoxin Chen, Wenyu Liu, Xinggang Wang, et~al.
\newblock Evf-sam: Early vision-language fusion for text-prompted segment anything model.
\newblock \emph{arXiv preprint arXiv:2406.20076}, 2024.

\bibitem[Zhu et~al.(2023)Zhu, Chen, Shen, Li, and Elhoseiny]{zhu_2023_minigpt}
Deyao Zhu, Jun Chen, Xiaoqian Shen, Xiang Li, and Mohamed Elhoseiny.
\newblock Minigpt-4: Enhancing vision-language understanding with advanced large language models.
\newblock \emph{arXiv preprint arXiv:2304.10592}, 2023.

\end{thebibliography}
}

\clearpage
\setcounter{page}{1}
\maketitlesupplementary

\section*{Supplementary}
In this supplementary material we discuss:
\begin{itemize}[noitemsep,topsep=1pt]
    \item the training protocol;
    \item further insights on the functioning of the Conditional Memory Encoder (CME), our learnable correction mechanism to adjust \sam{} tracking focus;
    \item additional ablations: on our Cross-Modal Temporal (CMT) Adapter, on inference window size, comparison with smaller backbones, and experiments on Referring Image Segmentation;
    \item comparison with \sam{}-based baselines;
    \item qualitative examples from MeViS to assess the effectiveness of \ours{} on challenging scenarios.
\end{itemize}

\section{Training protocol}
Following \cite{wu_2022_language}, we train our model with a  combination of DICE loss and binary mask focal loss.
We train our \textbf{Conditional Memory Encoder (CME)} via \textbf{self-supervision}. For each video clip, given the prompt $\rho$ we compute the predicted masks using \sam{} Mask Decoder:
\begin{equation}
    Y_m[t] = \mathcal{D}_{dec} (\mathcal{F}_{mem}, \rho) > 0, t=1..T.
\end{equation}
The predicted masks $Y_m[t]$ represent the standard output of \sam{} Mask Decoder, \ie the masks computed given the memory features $F_{mem}$.
As we aim at detecting when the \textit{memory-less} features highlight different object \wrt the one currently tracked, we further compute the unbiased output mask. By employing the unbiased \textit{memory-less} features, which do not take into account the previous tracking context encoded in the Memory Bank, the prediction is based solely on the object currently more aligned to the caption in the given clip. Formally:
\begin{equation}
    \mathcal{Y}_l[t] = \mathcal{D}_{dec} \left ( \mathcal{F}, \rho \right ) > 0, t=1..T.
\end{equation}
Given each pair of the binary masks at frame $t$, we define the detection label as: 
\begin{equation}
y_t =
\begin{cases}
  1
  & if \; \; \; \mathcal{Y}_l[t] \cap \mathcal{Y}_m[t] = 0 \\
  0
  & \; \; \; \; {otherwise}\\
\end{cases}
\end{equation}
The label is $1$ if the intersection of the two masks is null, \ie the masks segment different objects.
We supervise our CME with a standard Cross-Entropy loss:
\begin{equation}
\mathcal{L}_{CME} = - \frac{1}{T} \sum_{t=1}^{T}{[y_{t}log(p_{detect}) + (1 - y_{t})log(1-p_{detect})],}
\end{equation}
where $p_{detect}$ is computed as in eq. 9 of the main paper. 

\section{CME: Qualitative impact}
\begin{figure*}[t]
    \centering
    \includegraphics[width=1.0\linewidth]{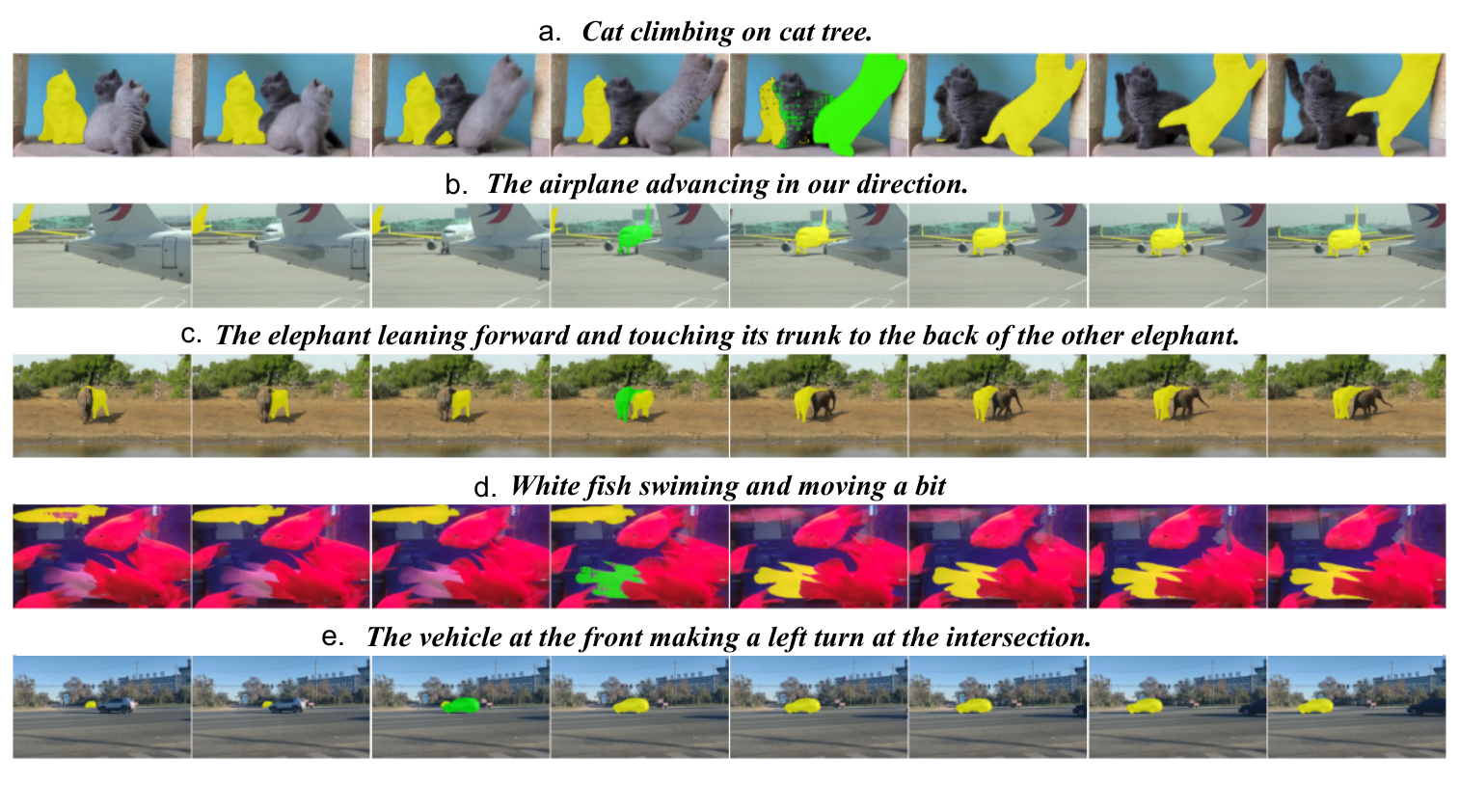}
    \vspace{-1.2cm}
    \caption{\textbf{Correct CME detections.} The plot shows examples where our CME correctly identifies (green masks) the referred object when the action starts unfolding. \ours{} recognizes that the newly proposed object is more aligned with the query and thus switches its tracking focus in the subsequent frames.}
    \label{fig:right_cme}
\end{figure*}

\begin{figure*}[t]
    \centering
    \includegraphics[width=1.0\linewidth]{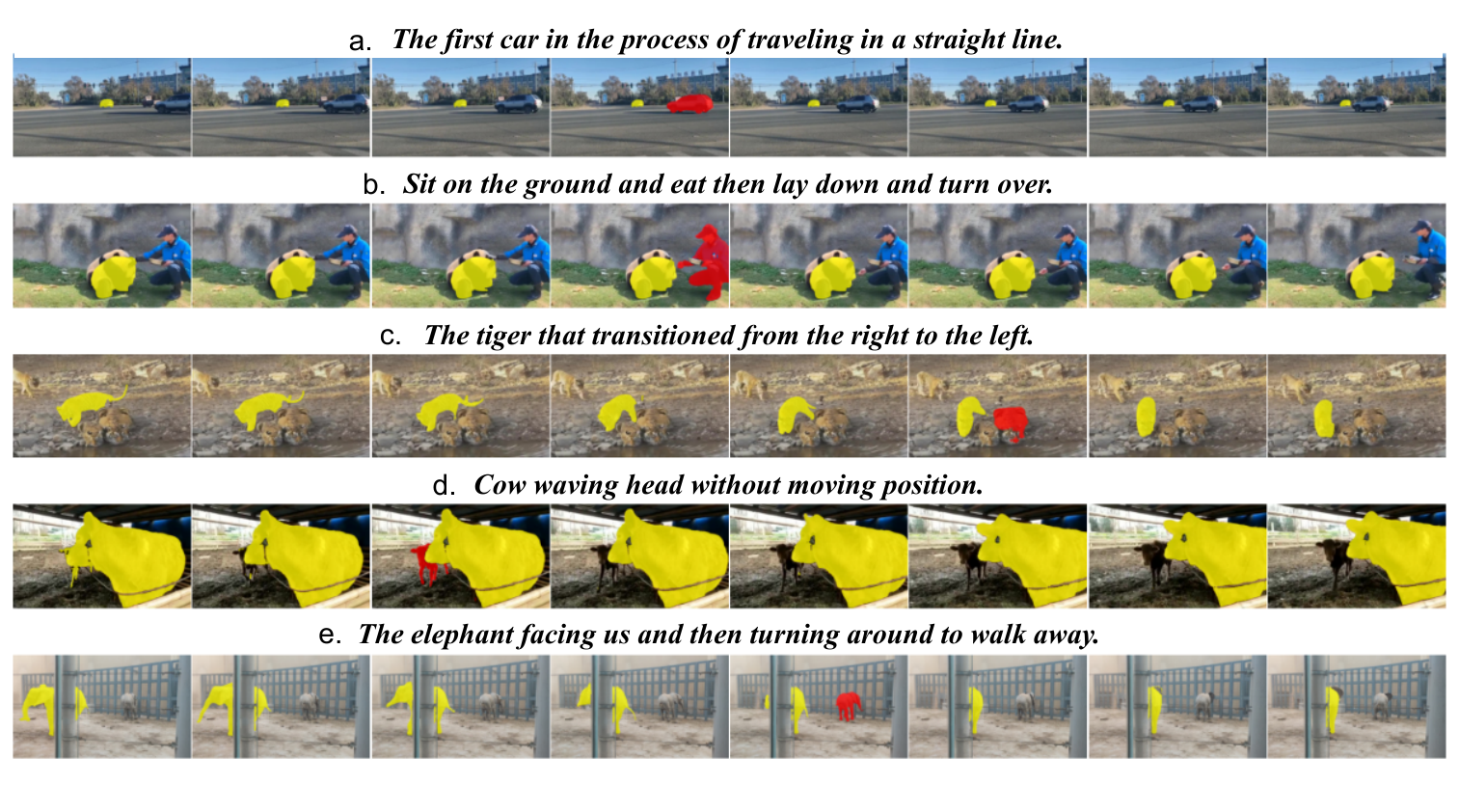}
    \vspace{-1.2cm}
    \caption{\textbf{Incorrect CME detections.}The plot shows examples where our CME provides wrong object proposals (red masks) due to lack of contextual information. In these examples, \ours{} determines that, when tacking into account past video context, the previously object is more aligned with the query and therefore does not switches its tracking focus. }
    \label{fig:wrong_cme}
\end{figure*}

In this section, we analyze the impact of the Conditional Memory Encoder (CME) within SAMWISE. In \cref{fig:right_cme} and \cref{fig:wrong_cme}, the model is tasked to segment the correct object in the video based on the provided referring expression. We use yellow masks to represent the output predictions generated by \ours{}.
Generally, the model tracks the object that appears most relevant according to the information available up to that point. However, due to the phenomenon of \textit{tracking bias}, \ie the tendency to continue tracking an initially detected object, the correct object might not be selected when it appears. Our CME addresses this challenge by detecting when an object aligned with the text prompt becomes visible. Upon detection, the CME computes the corresponding mask and encodes it into the Memory Bank. To highlight the CME role, we show the candidate masks it proposes in green or red, reflecting whether the proposed mask denotes a correct or incorrect detected object. For clarity, these masks are not predicted as final output but are temporary representations stored in the Memory Bank. By encoding these candidate masks, the CME enables \ours{} to adjust its tracking dynamically, balancing the influence of previously tracked objects with newly detected ones. 

\begin{figure*}[t]
    \centering
\includegraphics[width=1.0\linewidth]{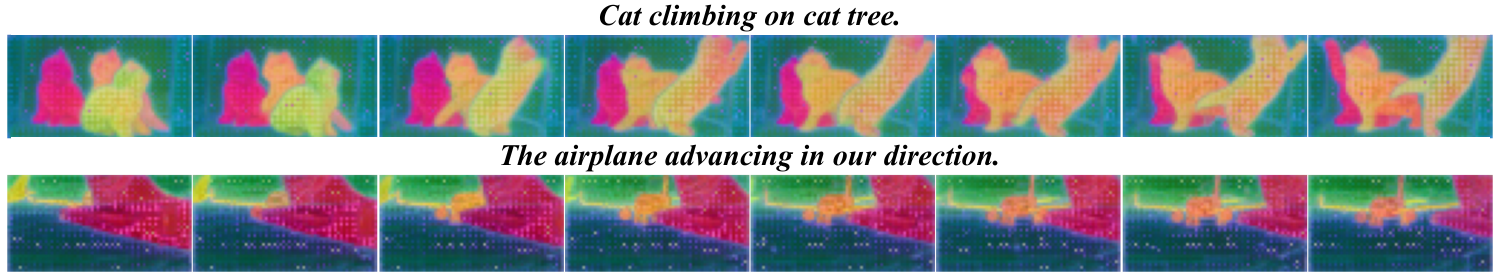}
    \caption{Effect of \textbf{Tracking bias.} The figure shows how \textit{memory features} (PCA) reinforce the initial choice, leading to tracking bias and preventing focus to more semantically aligned objects. In the first row, the model fails to shift attention when the correct object begins the relevant action; in the second, it misses the correct object when it appears later in the scene.}
    \label{fig:bias}
\end{figure*}

\myparagraph{Correct Object Detection by CME}
In \cref{fig:right_cme}, we showcase examples in which the CME successfully identifies the correct object. These examples highlight various challenging scenarios. In some cases, all potential objects are present in the scene from the beginning, but the discriminative action that distinguishes between them only occurs later in the video. For example, in case (a), the target cat starts \textit{climbing} only at a specific point in the sequence, and similarly, in case (c), the elephant \textit{touches its trunk to the back of the other elephant} at a later moment. In other scenarios, the action itself remains ambiguous until a key point. For instance, in example (e), the action of \textit{turning left} only becomes identifiable after a certain frame, at which point the CME detects the correct car and informs \ours{}, allowing it to shift focus to the correct instance. Similarly, in (d), the model faces a challenging scenario, where several instances are visible in the video and the action of \textit{moving a bit} remains ambiguous during the first frames. In other situations, like case (b), the target object is not visible at the start. Here, \ours{} starts tracking a different object (an incorrect airplane) until the target appears in the scene. 

\myparagraph{Handling Incorrect Candidate Detection}
In \cref{fig:wrong_cme}, we demonstrate the robustness of SAMWISE against incorrect candidate proposals generated by the CME. While our CME generates masks that align with the text prompt at clip-level, these proposals may not align correctly at a global level. This occurs because the CME reasons locally within the scope of the current clip, potentially leading to plausible but ultimately incorrect proposals.
Interestingly, SAMWISE is able to reason about past predictions and determine which object better aligns with the referring query, by relying on the broader context encoded in the Memory Bank. Therefore, the model is able to assess whether the candidate object is more aligned to the tracked object. We show this through a number of representative examples. 
For instance, in case (a), the CME proposes a novel plausible car (red mask). However, the previously tracked object was already \textit{traveling in a straight line}, and \ours{}, by balancing this contextual information with the new proposal, is able to correctly determine that the correct object is the one already subject to tracking.
Similarly, in case (d), the CME proposes a different cow, but \ours{} correctly interprets that \textit{waving head} describes more the foreground cow rather than the new one. In case (b), the referring expression is more ambiguous and lacks a specific subject, leading the CME to propose the human as the target object rather than the panda. However, \ours{} correctly identifies the panda as the object that aligns best with the query, as it is both \textit{sitting on the ground} and \textit{eating}. In example (e), the CME proposes the wrong elephant, but \ours{}, by reasoning over the frames, understands that the candidate object does not match the query, which describes an elephant \textit{turning around to walk away}.
Finally, in case (c), the described action has occurred in the past. The CME proposes a candidate tiger; however, \ours{}, by remembering which object actually \textit{transitioned from the right to the left}, refrains from switching its focus.

\section{Tracking Bias}
We provide additional qualitative examples to further exemplify the effect of \textit{tracking bias}, as visualized in \cref{fig:bias}, where we plot the \textit{memory features}.
Tracking bias occurs when the model mistakenly focuses on an incorrect object, failing to transition its attention to another, more relevant object once it emerges. This issue is particularly evident in scenarios where the target object becomes distinguishable only after performing a specific action. As shown in the examples, the model initial focus on an object causes it to overlook the presence of another, more semantically aligned instance, even when the latter matches the caption. This behavior stems from biased memory features, which reinforce the initial selection instead of adapting to new cues. 
\begin{table}
\begin{center}
\begin{adjustbox}{width=.8\linewidth}
\begin{tabular}{ccccccccc}
\toprule
\rowcolor{cyan!10} \multicolumn{5}{c}{Adapter layers} \\
Layer 1    & Layer 2 & Layer 3 & \multicolumn{1}{c|}{Params} & $\mathcal{J}$\&$\mathcal{F}$ \\
\midrule
& & & \multicolumn{1}{c|}{0.3 M} & 45.2 \\
           &            & \checkmark & 
\multicolumn{1}{c|}{2.2 M} & 50.3 \\
           & \checkmark & \checkmark & 
\multicolumn{1}{c|}{3.5 M} & 52.1   \\
\checkmark & \checkmark & \checkmark & 
\multicolumn{1}{c|}{4.2 M} & 54.2 \\
\midrule
\midrule
\rowcolor{cyan!10} \multicolumn{5}{c}{Hidden dimensionality} \\
\multicolumn{1}{c|}{} & 64 & 128 & 256 & 384 \\
\midrule
\multicolumn{1}{c|}{$\mathcal{J}$\&$\mathcal{F}$} & 48.0 & 52.1 &  54.2 & 52.5   \\
\multicolumn{1}{c|}{Params} & 1.0 M & 2.1 M & 4.2 M & 8.8 M \\
\bottomrule
\end{tabular}
\end{adjustbox}
\end{center}
%\vspace{-0.3cm}
\caption{Top: Ablation on the \textbf{Number of Adapters}. \textit{Layer i} indicates the intermediate layer of the Hiera backbone to which we add our CMT modules. Bottom: Effect of \textbf{hidden dimensionality} used inside our Cross-Modal Temporal Adapter.  
All numbers are reported without using our CME module, and CLIP-B as text encoder.}
\label{tab:layers}
%\vspace{-0.5cm}
\end{table}
\begin{table}{}{}
\begin{center}
%\noindent
\vspace{0.2cm}
\begin{adjustbox}{width=0.6\linewidth}
\begin{tabular}{c|cccccc}
\toprule
\rowcolor{cyan!10} Window & 4 & 6 & 8 & 12 \\
\midrule
$\mathcal{J}$\&$\mathcal{F}$ & 51.8 & 53.9 & 54.2 & 54.3  \\
\bottomrule
\end{tabular}
\end{adjustbox}
\end{center}
\vspace{-0.2cm}
\caption{\textbf{Effect of Window Size}. Ablation on the effect of window size (\ie number of frames processed together in each clip) in our online framework. Numbers combuted on MeViS \textit{valid-u} set, using CLIP-B as text encoder, without CME module.}
\vspace{0.2cm}
\label{tab:clip_len}
\end{table}
\begin{table}
\begin{center}
\begin{adjustbox}{width=\linewidth}
\setlength{\tabcolsep}{2pt}
\begin{tabular}{llcccccccccccccccccccccc}
\toprule
\multirow{2}{*}{Method} & Visual & Total  & MeViS & YT-VOS & DAVIS \\
\cline{4-6}  
& Encoder & Params & $\mathcal{J}$\&$\mathcal{F}$ &  $\mathcal{J}$\&$\mathcal{F}$ &$\mathcal{J}$\&$\mathcal{F}$  \\
\midrule
~~TCE-RVOS~\cite{Hu_2024_WACV}      \pub{WACV'24}  & ResNet-50  & - &
- &  59.6 & 59.4 \\
~~ReferFormer~\cite{wu_2022_language} \pub{CVPR'22} & ResNet-50  & 176 M &
-  & 58.7 & -  \\
~~OnlineRefer~\cite{wu_2023_online}   \pub{ICCV'23} & ResNet-50    & 176 M &
- & 59.3 & 57.3  \\
~~MUTR~\cite{yan2024referred}     \pub{AAAI'24} &  ResNet-50  & 190 M &
- &  61.9 & 65.3  \\
~~\ours~\scriptsize{(w/ CLIP-B)}                             & Hiera-B   & 150 M &
\underline{48.3} & \underline{67.2} & \underline{68.5}  \\
~~\ours                              & Hiera-B   & 202 M &
\textbf{49.5} & \textbf{69.2} & \textbf{70.6} \\
\bottomrule
\end{tabular}
\end{adjustbox}
\end{center}
\vspace{-0.3cm}
\caption{Comparison of \textbf{\ours{}} against state-of-the-art RVOS methods on MeViS, Ref-Youtube-VOS and Ref-DAVIS datasets using smaller backbones. \textbf{Bold} and \underline{underline} indicate the two top results.}
\label{tab:main_table_r50}
%\vspace{-0.4cm}
\end{table}

\section{Additional Ablations}
\myparagraph{Number of CMT adapters} In \cref{tab:layers}-top we assess how the number of adapters influences performance. Without any adapter (\ie relying only on a learnable MLP to project text prompts), the model achieves a modest $\mathcal{J}$\&$\mathcal{F}$ of 45.2\%. Adding a single adapter at the final layer, \ie on $\mathcal{F}^3$, provides a significant boost of $5.1\%$. Adding a second adapter, on $\mathcal{F}^2$, further improves performance by +$1.8\%$. Our chosen configuration, with three adapters across the last three layers of feature extractors, achieves the highest performance with a $\mathcal{J}$\&$\mathcal{F}$ of 54.2\%, indicating that multi-layer integration enhances feature refinement, thereby improving segmentation accuracy.

\myparagraph{Adapter hidden dimensionality}
In \cref{tab:layers}-bottom, we evaluate the performance of our CMT adapter with varying hidden dimensionalities. Our configuration, with a channel dimension of 256, achieves strong performance (54.2 $\mathcal{J}$\&$\mathcal{F}$) while maintaining a lightweight model with only 4.2M trainable parameters. Reducing the channel dimension to 64 or 128 results in a significant drop in performance, with a reduction in $\mathcal{J}$\&$\mathcal{F}$ of 6.2 and 2.1, respectively. Increasing the hidden dimensionality to 384 leads to a marginal performance drop of -1.7 $\mathcal{J}$\&$\mathcal{F}$, while doubling the number of trainable parameters (8.8 M).

\definecolor{mylightgray}{gray}{0.92}
\begin{table}[]
\begin{center}
\setlength{\tabcolsep}{3pt}
\begin{adjustbox}{width=\linewidth}
% \centering
\begin{tabular}{lcccccccccccccccccc}
\toprule
\multirow{2}{*}{Method} &  \multirow{2}{*}{\begin{tabular}[c]{@{}c@{}} Text \\ Encoder \end{tabular}}& \multicolumn{3}{c}{ \cellcolor{yellow!50} Referring Image Segmentation} \\
 &  & RefCOCO & RefCOCO+ & RefCOCOg    \\
\midrule
\rowcolor{mylightgray} \multicolumn{2}{l}{\emph{Large VLM based}} &&& \\
~VISA \pub{CVPR'24} & ChatUnivi & 72.4  & 59.8   & 65.5 \\
\rowcolor{mylightgray}
\multicolumn{5}{l}{\emph{RIS Specialist}} & \\
~MagNet \cite{chng2024mask} \pub{CVPR'24} & BERT & 75.2  & 66.2   & 65.4 \\
\midrule
%Referformer & RoBERTa & - & - & -  \\
\textbf{Ours} & RoBERTa  &
\textbf{76.8}  & \textbf{67.1}   & \textbf{67.3}  \\
\bottomrule
\end{tabular}
\end{adjustbox}
\end{center}
%\vspace{-0.6cm}
\caption{Comparison with SOTA for RIS. Results on the val set of the RefCOCO series dataset in terms of mIoU.}
\label{tab:sota_ris}
%\vspace{-0.4cm}
\end{table}

\begin{figure*}[h]
    \centering
    \includegraphics[width=1.0\linewidth]{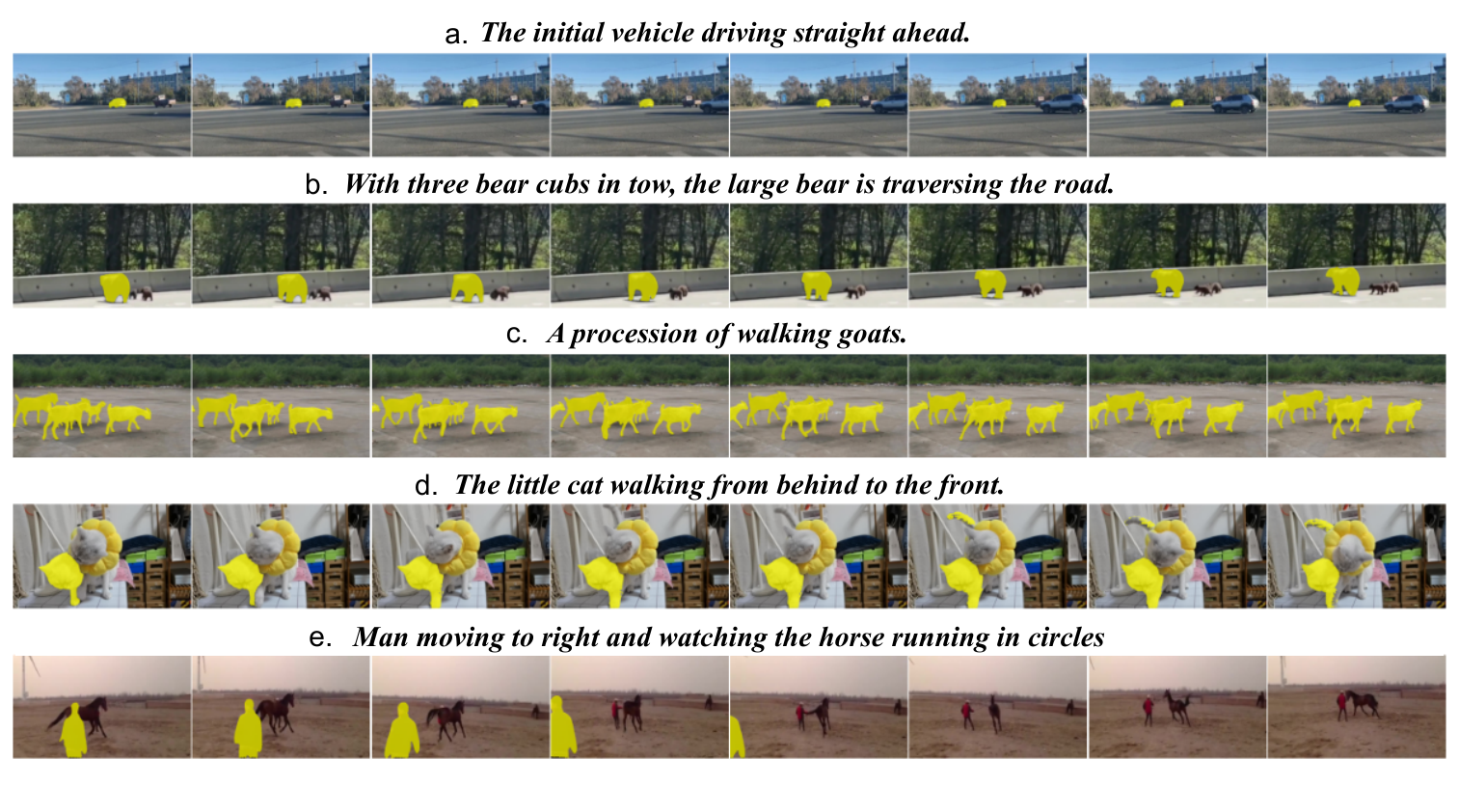}
    \vspace{-1.0cm}
    \caption{\textbf{Qualitative examples from MeViS.} The figure highlights SAMWISE ability to handle challenging RVOS scenarios, including occlusions, multiple instances, and distinguishing between similar objects based on actions and descriptive attributes.}
    \label{fig:neutral}
\end{figure*}
\myparagraph{Window size} In \cref{tab:clip_len} we evaluate how the number of frames in each processed clip affects performances. Performances increase with the number of frames, as a larger window allows to better model temporal evolution. Since increasing the window size from 8 to 12 only yields marginal gains, we chose to keep 8 as clip length to better suit an online framework.

\myparagraph{Comparison with smaller backbones} In \cref{tab:main_table_r50} we compare against previous methods using a smaller backbone, namely a ResNet-50. In this setting we obtain comparable model sizes and higher perfoemance gap.

\myparagraph{Referring Image Segmentation} Among our contributions, the design of the HSA and the CME module are tailored to address challenges of referring segmentation in videos. However, the fundamental value of our CMT adapter is that it enables prompting \sam{} with referring expressions, which can be thus easily applied for image-level tasks. In \cref{tab:sota_ris} we evaluate \ours{} on Referring Image Segmentation benchmarks, comparing against state-of-the-art specialist models, and Large-VLM based. Remarkably, we find that our \ours{} achieves competitive results also in image-level tasks, showcasing its versatility.

\begin{table}
\begin{center}
\begin{adjustbox}{width=\linewidth}
% \centering
\begin{tabular}{lcccccccccccccccccc}
\toprule
\multirow{2}{*}{Method} & MeViS & Ref-YT-VOS & Ref-DAVIS \\
\cline{2-4}
&  $\mathcal{J}$\&$\mathcal{F}$ & $\mathcal{J}$\&$\mathcal{F}$ & $\mathcal{J}$\&$\mathcal{F}$  \\
\midrule
~~G.DINO+SAM2 \textit{1st frame}      &
37.7  & 57.5  & 66.4  \\
~~G.DINO+SAM2 \textit{All frames}     &
36.8 & 56.9 & 61.2 \\
~~\ours~\textbf{(ours)}               &
\textbf{48.3}  & \textbf{67.2} & \textbf{68.5} \\
\bottomrule
\end{tabular}
\end{adjustbox}
\end{center}
\vspace{-0.2cm}
\caption{Comparison of \textbf{\ours{}} against baselines that employ an off-the-shelf grounded detector (GroundingDino) to provide box prompts.
}
\label{tab:dino}
%\vspace{-0.2cm}
\end{table}

\myparagraph{Training time}
Training our pipeline requires roughly 150 GB of GPU memory. In our setup, this translates in training on 2 A100 for 18 hours for finetuning on MeViS. Full fine-tuning of SAM2, \ie without our adapters, requires roughly 3 times more GPU memory. For the experiment on full-finetuning (Tab. 4 of main paper), we used 8 A100 for 26 hours.

\section{SAMWISE vs naive baselines with SAM2}

In \cref{tab:dino}, we compare SAMWISE with two baselines utilizing SAM2:
\begin{itemize}
    \item \textbf{GroundingDINO + SAM2 \textit{1st frame}}: This approach employs GroundingDINO \cite{liu_2023_grounding} to identify the referred object in the first frame based on the textual query. The resulting bounding box is then used to prompt SAM2 \cite{ravi_2024_sam2}, which tracks the object across the video.
    \item \textbf{GroundingDINO + SAM2 \textit{All frames}}: In this baseline, GroundingDINO \cite{liu_2023_grounding} detects the referred object in each frame using the textual query. The bounding box is then used to prompt SAM2 \cite{ravi_2024_sam2} independently on each frame.
\end{itemize}

\noindent
Results indicate that \ours{}{} consistently outperforms both baselines. Specifically, it surpasses them by approximately 10\% in $\mathcal{J}$\&$\mathcal{F}$ on both MeViS \cite{ding_2023_mevis} and Ref-Youtube-VOS \cite{bellver_2023_refvos}, and by 2\% and 7\% on Ref-DAVIS \cite{khoreva_2019_davis}, respectively. 
GroundingDINO + SAM \textit{1st Frame} baseline heavily relies on the accuracy of the initial bounding box proposal since the object is identified solely in the first frame and then tracked. This dependency leads to suboptimal results, especially when the target object cannot be clearly identified in the first frame, either because the object appears later or the relevant action unfolds as the video progresses. However, this baseline performs relatively well on Ref-DAVIS \cite{khoreva_2019_davis}, which contains more static, object-centric videos. The second row shows the results for GroundingDINO + SAM \textit{All Frames}. Although this method allows for frame-by-frame object detection, it does not leverage \sam{} tracking capabilities, leading to poor masks quality. 
Additionally, limiting reasoning to individual frames causes the model to overlook temporal consistency, often resulting in shifts between objects across frames. 
In contrast, \ours{} explicitly models temporal evolution within its features and integrates textual cues without relying on external bounding box proposals. This design enables consistent localization, segmentation, and tracking of the target object.

\section{Qualitative results}
In \cref{fig:neutral}, we present qualitative examples from the MeViS dataset that highlight the effectiveness of \ours{}. These examples cover a range of challenges typical in RVOS. \ours{} shows strong robustness in dealing with occlusions (case e.), accurately tracking target objects even when they are partially or fully obscured. It also handles situations with multiple instances (case c.), correctly segmenting all relevant objects.
Additionally, \ours{} excels at disambiguating between similar objects by reasoning over both actions (cases a. and b.) and descriptive attributes (case b.), ensuring precise identification of the correct targets based on their behavior and characteristics in the scene.

\end{document}